\documentclass[11pt]{article}

\usepackage{arxiv}

\usepackage[T1]{fontenc}    % use 8-bit T1 fonts
\usepackage{url}            % simple URL typesetting
\usepackage{nicefrac}       % compact symbols for 1/2, etc.
\usepackage{microtype}      % microtypography
\usepackage{doi}
\usepackage[square,sort,comma,numbers]{natbib}
\usepackage[title]{appendix}
\usepackage{hyperref}
\usepackage[utf8]{inputenc}
\usepackage{booktabs} % for borders and merged ranges
\usepackage{multirow}
\usepackage{soul}% for underlines
\usepackage[table]{xcolor} % for cell colors
\usepackage{changepage,threeparttable} % for wide tables
\usepackage{caption}
\usepackage{subcaption}
\usepackage{graphicx}
\usepackage{amsmath}
\usepackage{float}
\usepackage{authblk}
\usepackage{amsmath,amsfonts,amssymb}
\usepackage{algpseudocode}
\usepackage{fancyvrb}
\usepackage{bm}   
\usepackage[pagewise]{lineno}

\usepackage{makecell}

\usepackage[resetlabels]{multibib}
\newcites{supp}{Supplementary References}

\setstcolor{red}
\usepackage[normalem]{ulem}
\newcommand\redsout{\bgroup\markoverwith{\textcolor{red}{\rule[0.5ex]{2pt}{0.4pt}}}\ULon}

\newcommand{\beginsupplement}{%
    
        \setcounter{table}{0}
        \renewcommand{\thetable}{S\arabic{table}}%
        \renewcommand{\theHtable}{Supplement.\thetable}
        \setcounter{figure}{0}
        \renewcommand{\thefigure}{S\arabic{figure}}%
        \renewcommand{\theHfigure}{Supplement.\thefigure}
        \setcounter{section}{0}
        \renewcommand{\thesection}{S\arabic{section}}%
        \renewcommand{\theHsection}{Supplement.\thesection}
}

\DeclareMathOperator*{\argmin}{arg\,min} % Jan Hlavacek

\author[, 1]{Maxime Delmas\thanks{Corresponding author: maxime.delmas@idiap.ch}}
\author[2]{Magdalena Wysocka}
\author[1,2,3]{Andr\'{e} Freitas}
\affil[1]{Idiap Research Institute, Switzerland}
\affil[2]{Digital Experimental Cancer Medicine Team, Cancer Biomarker Centre, CRUK Manchester Institute}
\affil[3]{Department of Computer Science, University of Manchester}
\date{}
\title{Relation Extraction in underexplored biomedical domains: A diversity-optimised sampling and synthetic data generation approach}

\begin{document}

% --- pour les captions des tables ??
% \captionsetup{skip=0.5\baselineskip}

\maketitle

%------------------------------

\section*{Abstract}

% Natural products represent a large pool of bioactive compounds of high interest in drug-discovery. The LOTUS database has significantly enhanced the accessibility, sharing, and management of organism-chemical relationships reported in the scientific literature. However, these relationships are sparsely distributed across organisms and a growing part of the literature remains unannotated. This volume necessitates the development of a machine assistant to boost the completion of existing resources. 
The sparsity of labelled data is an obstacle to the development of Relation Extraction models and the completion of databases in various biomedical areas. While being of high interest in drug-discovery, the natural-products literature, reporting the identification of potential bioactive compounds from organisms, is a concrete example of such an overlooked topic. To mark the start of this new task, we created the first curated evaluation dataset and extracted literature items from the LOTUS database to build training sets. To this end, we developed a new sampler inspired by diversity metrics in ecology, named Greedy Maximum Entropy sampler, or GME-sampler (\url{https://github.com/idiap/gme-sampler}). The strategic optimization of both balance and diversity of the selected items in the evaluation set is important given the resource-intensive nature of manual curation. After quantifying the noise in the training set, in the form of discrepancies between the input abstracts text and the expected output labels, we explored different strategies accordingly. Framing the task as an end-to-end Relation Extraction, we evaluated the performance of standard fine-tuning with models such as BioGPT, GPT-2 and Seq2Rel, as a generative task, and few-shot learning with open Large Language Models (LLaMa 7B-65B). Interestingly, the training sets built with the GME-sampler also exhibit a tendency to tip the precision-recall trade-off of trained models in favour of recall. In addition to their evaluation in few-shot settings, we explore the potential of open Large Language Models (Vicuna-13B) as synthetic data generator and propose a new workflow for this purpose. All evaluated models exhibited substantial improvements when fine-tuned on synthetic abstracts rather than the original noisy data. We provide our best performing ($\text{f1-score}=59.0$) BioGPT-Large model for end-to-end RE of natural-products relationships along with all the generated synthetic data and the evaluation dataset. See more details at \url{https://github.com/idiap/abroad-re}.

% si jamais c'est trop long: on retire :  Interestingly, the training sets built with the GME-sampler also exhibit a tendency to tip the precision-recall trade-off of trained models in favour of recall.

% et on remplace: Framing the task as an end-to-end Relation Extraction, we evaluated the performance of standard fine-tuning with models such as BioGPT, GPT-2 and Seq2Rel, as a generative task, and few-shot learning with open Large Language Models (LLaMa 7B-65B). Par:  Framing the task as an end-to-end Relation Extraction, we evaluated the performance of standard fine-tuning and few-shot learning. 
%------------------------------

\section{Introduction}
% CONTEXT
%% Big problem

The biomedical literature constitutes a vast but still underexploited reservoir of knowledge, the growth of which reflects the expansion of topics and areas of applications. However, the diversity and morphological richness of bio-entities and the complexity of the relationships expressed between them, contrast with the sparsity of the available labelled data. While some domains can already benefit from efficient extraction models (e.g chemical-disease relationships) for database completion, less popular domains, like the natural product (NP) literature, are often overlooked. NPs are chemical compounds produced by living organisms (plants, bacteria, fungi, etc.), exhibiting a wide range of structure and functions and offering a vast reservoir of potential therapeutic molecules. The isolation and identification of NPs is primarily reported in the scientific literature and also disseminated in different public databases (eg. COCONUT \cite{sorokina_coconut_2021}, KNApSAck \cite{shinbo_knapsack_2006}, etc.). Recently, the LOTUS initiative \cite{rutz_lotus_2022} has successfully established an Open and FAIR standard resource for natural product chemistry through a rigorous harmonization of a heterogenous set of databases. However, the extent of the NP landscape is not reflected by the content of the databases, which are incomplete and exhibit an imbalanced coverage toward model organisms (eg. \textit{A. Thaliana}). While a significant portion of the existing literature remains unannotated, there is also a continuous surge of new publications reporting novel relationships that could contribute to filling this gap.

Enriching such knowledge bases requires jointly performing Named Entity Recognition (NER) and Relation Extraction (RE). In this case, NER is defined as a sub-task which consists of identifying the boundaries and classifying the type of named entities (i.e an organism "Isaria sinclairii" and a chemical "\textit{fingolimod}"\footnote{\url{https://lotus.naturalproducts.net/compound/lotus_id/LTS0203935}}). The subsequent RE step is the semantic classification of the relations between two (or more) entities. To complete NP databases, the objective is to extract the "\textit{produces}" or "\textit{is isolated from}" relationships between organisms and chemicals. Noteworthy, others types of relationships can also be expressed, such as "\textit{inhibits the growth of}". Traditional deep learning models exhibiting SOTA performance on NER and RE (separately or in so-called \textit{end-to-end} models) relies on a large set of labelled data \cite{luo_biogpt_2022, giorgi_sequence--sequence_2022, wang_global--local_2020}. However, while datasets like Linneaus \cite{gerner_linnaeus_2010} have been successfully applied for organisms recognition, existing chemical NER datasets (i.e CHEMDNER \cite{krallinger_chemdner_2015}) do not provide sufficient coverage on the NP literature and do not adequately capture their morphological specificities. Along with the typically long systematic names of metabolites (eg. \texttt{3'-[gamma-hydroxymethyl-(E)-gamma-methylallyl]-2,4,2',4'-tetrahydroxychalcone 11'-O-coumarate}\footnote{\texttt{PMID: 11678652}}), many chemical mentions are defined as Multiples co-joined enumerations, where entities are mentioned in non-continuous strings such as "cystodiones A-D" or "wortmannins C and D", and are particularly frequent. These chemical mentions must be correctly identified and expanded to recover the full list of entities, which also adds complexity to the decoding process. Finally, to the best of our knowledge, no datasets are available for the subsequent RE step \cite{luo_biored_2022}. The aforementioned constraints are frequently encountered in BioNLP, when venturing beyond the well-studied chemical-disease associations or protein-protein interactions. 

Meanwhile, the abundance of unlabelled textual data has been instrumental in driving recent breakthroughs in representation learning \cite{wysocki-etal-2023-transformers} and the development of the foundational models (e.g GPT and LLaMA model families). The zero/few-shot learning capabilities of Large Language Models (LLMs) \cite{kojima_large_nodate, brown_language_nodate} make them serious candidates for performing a task with only a handful of examples. Moreover, conversations (chatbots) and instructions-tuned models \cite{zhang_instruction_2023} also represent a promising opportunity for synthetic data generation to alleviate the main problem, namely the lack of labelled data within the target distribution. Indeed, beyond the sophistication of model architectures, data availability and quality are limiting factors for the extraction performance, but often neglected \cite{sambasivan_everyone_2021}. 

\begin{figure*}
\begin{center}
  \includegraphics[width=.2\textwidth]{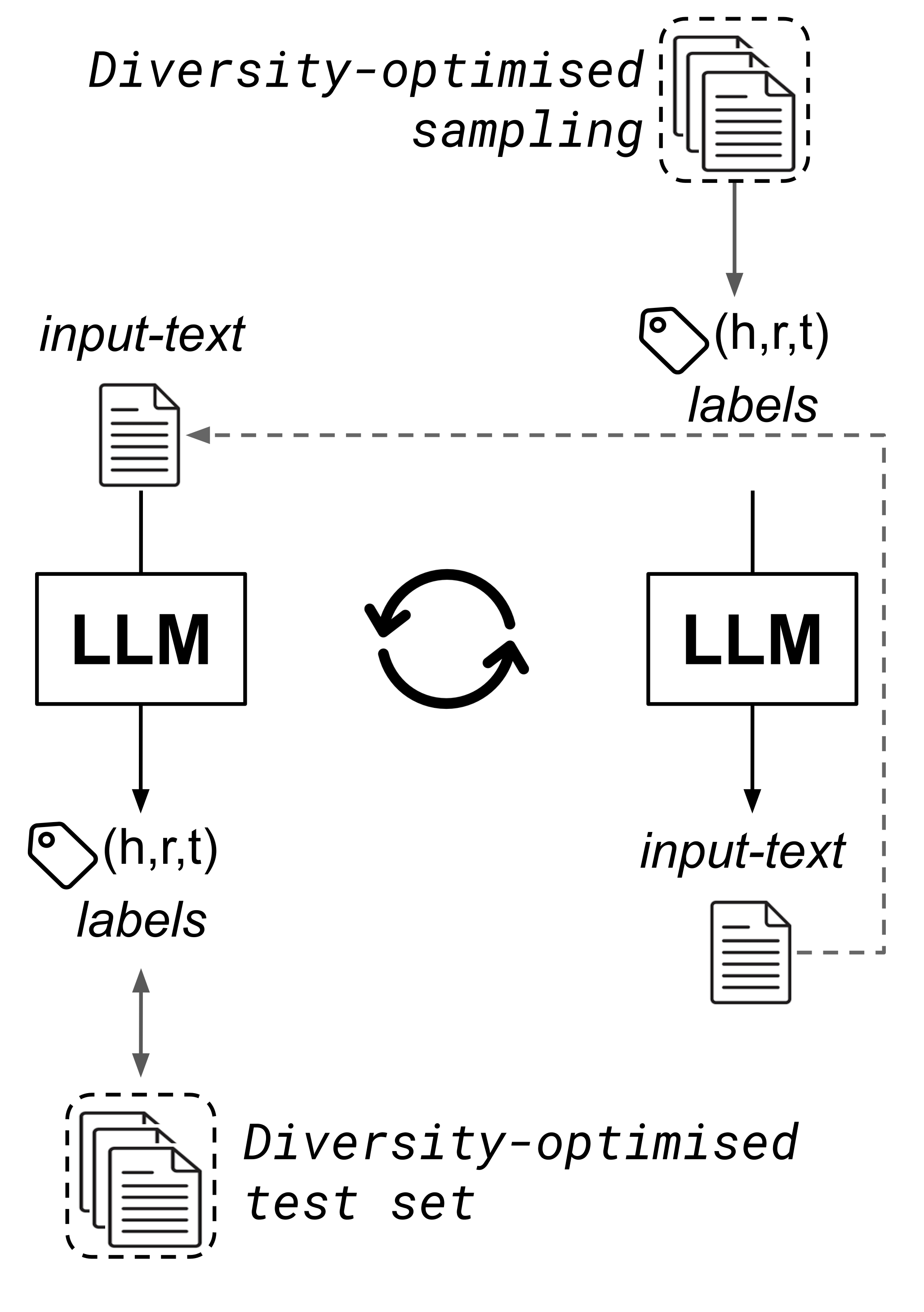}
\end{center}
\caption{Combining diversity-optimised sampling and synthetic data generation.}
\label{fig:schemahypothesis}
\end{figure*}

In order to address these scarcity constraints, we propose an end-to-end generative extraction paradigm, which introduces two novel methodological contributions. Firstly, we introduce a diversity-optimised sampling strategy, which minimises the selection of items for the parsimonious creation of evaluation gold-standards and training sets. This component minimises the popularity biases and associated imbalance towards entities which are over-expressed in the literature (e.g. model organisms and recurring substances), allowing for a systematic (entropy-based) method to maximise diversity and measure the utility of new annotations. Secondly, we use the generative expressivity of models fine-tuned on conversations and instructions for creating within distribution synthetic data, to support the construction of end-to-end joint NER-RE extraction models. In this framework, the diversity sampled entities and associated relations are linguistically embedded within synthetically generated text. The overall framework is depicted in Figure \ref{fig:schemahypothesis}.

%The imposed constraints have thus motivated the application of generative models in an end-to-end framework and the evaluation of alternative strategies with LLMs using sparse or poorly labelled data. As illustrated in Figure \ref{fig:schemahypothesis}, we evaluated the following research hypotheses from developing a dataset to comparing different strategies:

More formally, this paper aims to investigate the following research hypotheses (RHs) as supporting mechanisms for addressing these limitations, using NPs as a validation domain:

\begin{itemize}
    \item \textbf{RH1: Diversity-optimised sampling provide a valuable selection of items to build training and evaluation datasets for RE.}
    \item \textbf{RH2: In a practical scenario with noisy labels, LLMs can be more beneficial as synthetic data generator than unsupervised predictors.}
\end{itemize}

\begin{figure*}
\begin{center}
  \includegraphics[width=.8\textwidth]{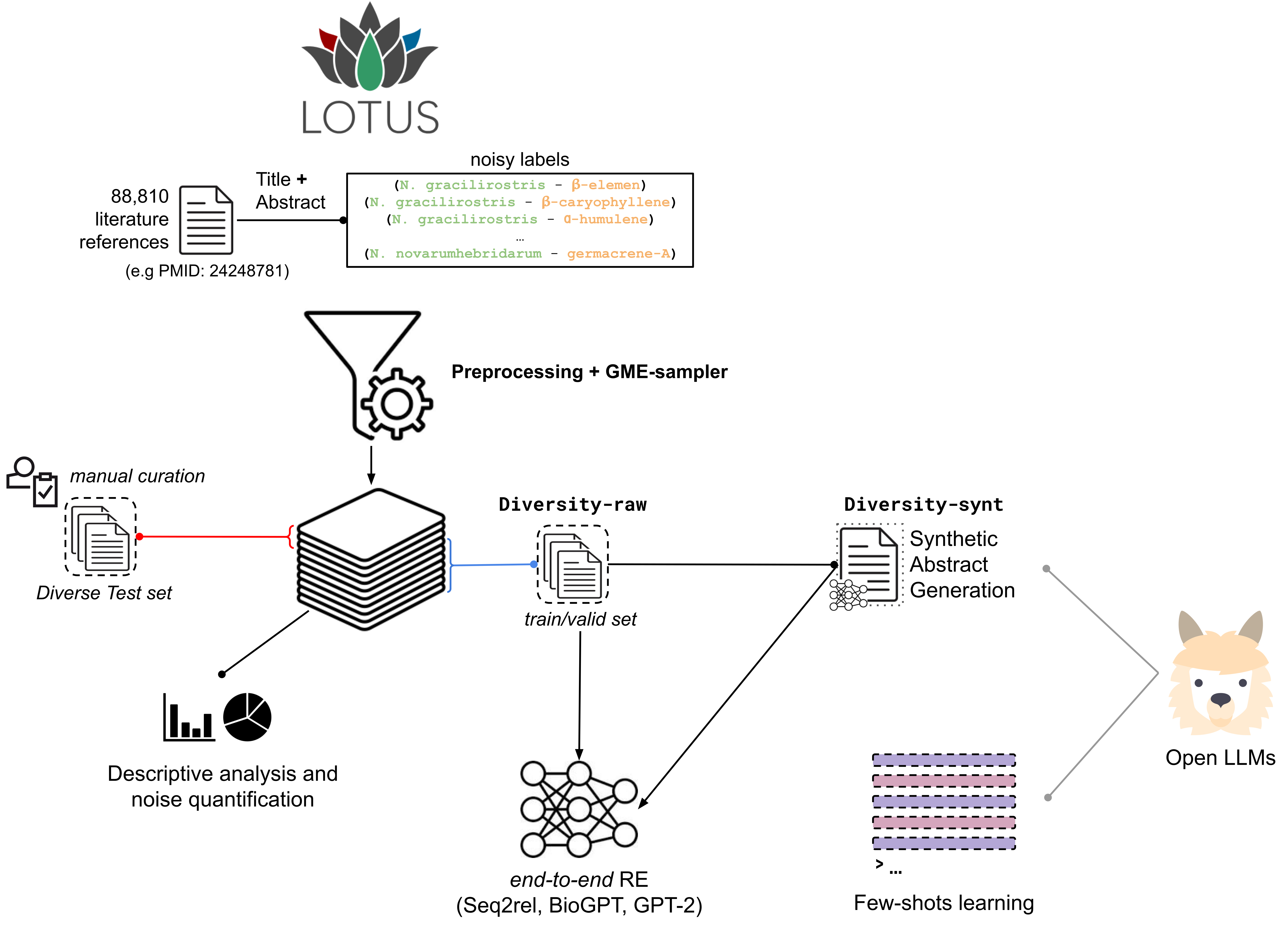}
\end{center}
\caption{A global diagram of the workflow presented in this work.}
\label{fig:schemaWork}
\end{figure*}

With the aim of providing an end-to-end RE model to help expanding natural products databases, we started by building a training and evaluation dataset. Inspired by the metrics used in ecology, we first proposed the Greedy-Maximum-Entropy sampler (GME-sampler) to extract a diversity-optimised sample from the LOTUS database. By manually annotating the top-diverse items, we proposed the first evaluation dataset for this task, which can serve as a benchmark for future developments in this area. Following a descriptive analysis of the remaining data and quantifying the noise present in the form of discrepancies between raw input text and annotated (standardized) labels, we evaluated various modelling approaches. First, we compared the performance of standard fine-tuning techniques on the available noisy data to the few-shot learning capabilities of open LLMs. Leveraging the generative capabilities of a LLM (Vicuna-13B), we then proposed a novel synthetic abstract generation pipeline and demonstrated the significant performance improvements (on average $24.7\%$ in $\text{f1-score}$) brought by these new training data on the evaluated models. In line with these results, we have made available our best-performing BioGPT-Large model ($\text{f1-score}=59.0$) and the $\approx 25000$ synthetic abstracts on which it has been trained. A synthetic diagram of the different strategies explored in this work is presented in Figure \ref{fig:schemaWork}. The main contributions of the work are summarised below:
\begin{itemize}
    \item A diversity-optimised sampler (GME-sampler) for building diverse and balanced datasets for RE (see \url{https://github.com/idiap/gme-sampler}).
    \item The first curated evaluation dataset for RE between organisms and natural-products (see \url{https://zenodo.org/records/8422007}).
    \item An evaluation of different strategies for RE with noisy labels.
    \item A framework for synthetic data generation via chatbot or instruction-tuned models and produced training datasets (see \url{https://github.com/idiap/abroad-re} and \url{https://zenodo.org/records/8422294}).
    \item A set of ready-to-use BioGPT fine-tuned models (see \url{https://huggingface.co/mdelmas/BioGPT-Large-Natural-Products-RE-Diversity-synt-v1.0})
\end{itemize}

% Related works
\subsection{Related work}

Biomedical RE \cite{shahab_short_2017, zhao_recent_2020, zhao_comprehensive_2023} encompasses various subtypes, depending on the considered bio-entities, such as drug-drug interactions \cite{zhang_deep_2020}, chemical-disease relationships \cite{li_biocreative_2016}, gene-disease associations \cite{su_renet2_2021}, and protein-protein interactions \cite{ahmed_identifying_2019}, among the most popular. Investigating the overlooked NPs relationships necessitated the exploration of several interconnected sub-tasks, including the selection and partitioning of a dataset, the generation of synthetic data and the assessment of various end-to-end RE strategies. This section provides a review of the closely-related works that align with these three development axes.

\subsubsection{Splitting datasets and impact of diversity}

Data selection and partitioning methods can significantly impact the generalization performance of supervised models. \citet{xu_splitting_2018} evaluated various splitting techniques, including K-S \cite{kennard_computer_1969} and SPXY \cite{galvao_method_2005}, and emphasized the importance of maintaining a balance between training and test sets for a reliable evaluation of models. Like the recently proposed SPlit \cite{joseph_split_2022} method, these approaches aim to select a representative subset of the data, leveraging different distance metrics. Unlike the Euclidean or Energy-based distances used in aforementioned methods, the GME-sampler uses an entropy-based metric to capture diversity and select representative evaluation and training sets.
Regarding diversity, \citet{yu_can_nodate} investigated various diversity-based metrics for selecting training data, and demonstrated their positive impact on the performance of NER models. Additionally, other works have highlighted the significance of effective data selection over a naive increase of the dataset size for training \cite{axelrod_domain_nodate, fan_learning_2017, feng_reinforcement_2018}.

\subsubsection{Synthetic data generation}

Training neural RE models strongly rely on a substantial and diverse set of training data. However, annotating large datasets with experts is time-consuming and costly. To overcome this limitation, many works explored approaches such as Data-augmentation (DA) \cite{hu_gda_2023, feng_survey_2021, pellicer_data_2023} and Distant Supervision (DS) \cite{smirnova_relation_2018, mintz_distant_2009}, which enable the expansion of the dataset size by creating new training examples from existing ones, or, by assigning pseudo-labels to external, unlabelled data. In the biomedical domain, the RE challenge ChemProt \cite{yoon_biomedical_2023, iinuma_improving_2022} or Protein-Protein interactions extraction \cite{su_using_2019}, have recently benefited from the application of these methods. Synthetic data generation (SDG) goes beyond DA or DS by creating fully synthetic datasets, i.e paired input text and output labels. A significant body of influential works has leveraged the generative capabilities of LLMs to propose different SDG strategies in zero-shot \cite{ye_zerogen_2022, gao_self-guided_2023, schick_generating_2021, he_generate_2022, wang_towards_2021, smith_language_2022, meng_generating_2022, kumar_data_nodate}, few-shot \cite{bonifacio_inpars_2022, dai_promptagator_2022, meng_tuning_2023, chen_weakly_2022, yoo_gpt3mix_2021}, or by fine-tuning \cite{anaby-tavor_not_2019, papanikolaou_dare_2020, hartvigsen_toxigen_2022}. Similarly to this work, \citet{josifoski_exploiting_2023} also proposed to reverse the task and used LLMs from OpenAI to generate plausible input text based on expected output triplets from Wikidata. \citet{tang_does_2023} compared the performance of LLM (ChatGPT) in directly extracting information from unstructured clinical text to its potential use as synthetic data generator for DA. \citet{veselovsky_generating_2023} evaluated various prompting strategies to improve diversity and alignment between synthetic and real-world data distributions for sarcasm detection. \citet{yang_generative_2020} combined synthetic data generation with a diversity-augmentation component for common sense reasoning. \citet{aggarwal_ecg-qalm_2023} applied SDG to biomedical NER, while \citet{xu_s2ynre_2023} used a two-stage training procedure on synthetic and golden data, notably for extracting protein interactions with the ChemProt dataset. In contrast, this work proposes to leverage Open LLMs to generate synthetic abstracts based on a list of verbalised main findings. The diversity of the generations is increased and guided by the entropy-based sampling of the seed articles which originally report these findings, as well as a set of crafted patterns of expressions.

\subsubsection{End-to-End Relation Extraction}

\citet{kambar_survey_2022} classifies various strategies and highlights the potential of end-to-end (or joint) NER and RE methods to overcome limitations of the traditional pipeline approaches. In the biomedical domain, \citet{li_neural_2017} proposed a Bi-LSTM for drug adverse effects extraction, while \citet{esmail_zadeh_nojoo_kambar_chemical-gene_2022} introduced a GNN for chemical-protein interactions. Recent approaches frames the task in a generative "text-to-text" process, using sequence-to-sequence models, by lineralising the expected relations as a text string to be decoded from the input. Seq2Rel \cite{giorgi_sequence--sequence_2022} and REBEL \cite{huguet_cabot_rebel_2021} proposed different linearisation schemas, and \citet{zhang_minimize_2020, zeng_learning_2019} notably assessed the biases caused by the forced order of relationships during training. \citet{hou_discovering_2021} trained a sequence-to-sequence model for drug-target interactions extraction, and \citet{zeng_extracting_2018} introduced a copy mechanism. Additionally, \citet{eberts_end--end_2021-1} used four task-specific sub-components, and \citet{paolini_structured_2021} utilized a translation mechanism. Finally, BioGPT \cite{luo_biogpt_2022} demonstrated SOTA performance on several biomedical datasets using an autoregressive approach, providing the input text as context.

\section{Proposed approach}
This section describes the different methodology used in this work. We start by describing our first contribution, the GME-sampler, in section \ref{sec:GME-meth}. The few-shot learning and fine-tuning strategies evaluated for the RE task are then described in sections \ref{meth:icl} and \ref{meth:ft}. The synthetic abstract generation procedure is described in \ref{sec:meth-synth-generation}. Finally, details on the evaluation, experimental setup and implementation details are provided in supplementary \ref{subsec:setup-and-details}.

\subsection{Greedy Maximum Entropy Sampling (GME)}
\label{sec:GME-meth}
The objective is to extract a sample $S$ of documents from an initial set $D$ with an optimized diversity of mentioned organisms and chemicals: $S \subset D$, $|S| = l$ and $|D| = L$. The initial set $D$ correspond to the LOTUS dataset, in which each document $d$ reports a set of relations between organism(s) and isolated natural product(s): $d = \{r_1, r_2, ..., r_{n_d}\}$, where $n_d$ is the number of reported relations in $d$. A relation $r_k = (o_i, c_j)$ involves the organism $o_i$ and the chemical $c_j$. The set of organisms and chemicals are denoted as $O$ and $C$, respectively.\\
Then, given a set $S$ of documents, the probability that a reported relation involve the organism $o_i$ (and similarly for the chemical $c_j$) is 
\begin{equation}
P(o_i) = \frac{\displaystyle\sum_{d \in S} |\{ r_k: o_i \in r_k \}^{d}|}{\displaystyle\sum_{d \in S} n_d}, \text{ with } |\{r_k: o_i \in r_k \}^{d}| \text{ the number of relations involving $o_i$ in $d$.}
\end{equation}
It follows that the diversity of organisms or chemicals can be measured with the Shannon's entropy over the probability distributions of elements of $O$ and $C$ in the sample $S$. Expressed with entropy, the diversity reflects the uncertainty about the organism, or the chemical, which is attached to a relation reported in an article \cite{leinster_entropy_2022, jost_entropy_2006}. For organisms (resp. chemicals) the Shannon's entropy in the sample $S$ is $H_S(O) = - \displaystyle\sum_{o_i \in O} P(o_i) \log P(o_i)$. Adding a new document $\displaystyle d$ to $S$ will update the probability distributions of $O$ (resp. $C$), and the new observed entropy will be $H_{\displaystyle S_{+d}}(O)$ where $\displaystyle S_{+d} = S \cup \{d\}$. Therefore, to optimize the diversity over organisms and chemicals, the document $d*$ added to $S$ minimizes the distance to the utopian point $(log |O|, log |C|)$ (maximal observable entropy over organisms and chemicals):
\begin{equation}\label{eq:utopian}
d* = \argmin_d || (H_{\displaystyle S_{+d}}(O), H_{\displaystyle S_{+d}}(C)) - (log |O|, log |C|) ||.
\end{equation}
The proposed sampling approach is a simple greedy algorithm that, at each step, selects and adds the new document $d*$ from $D$, maximising the diversity on organisms and chemicals (see eq.\ref{eq:utopian}). We refer to it as diversity-sampling. As in ecology, diversity is intrinsically related to the number of distinct biological entities (e.g organisms) \cite{hill_diversity_1973}, but, for the purpose of using the samples as training or evaluation sets, the balance of the distribution is prioritized over a higher number of rare entities \cite{leinster_entropy_2022}. This latter behavior is a natural consequence of using the Shannon's entropy. From this perspective, the method can also be seen as a ranking procedure, and a sample is determined by selecting the first top $n$ ranked items. The selection of an appropriate sample size $l$ is also a critical, but often overlooked factor. By monitoring $H_S(O)$ and $H_S(C)$ during the iterative construction of $S$ (until $l=L$), it is possible to determine the step $l$ at which diversity starts to deteriorate and sampling should be stopped, i.e. when the new added documents provide relationships for already frequently reported entities in $S$. The GME-sampler initially designed for the purpose of extracting data from LOTUS has also been implemented as a standalone library and is readily applicable to alternative contexts with \texttt{N-ary} relations (e.g Pharmacogenomics: \textit{Variant - Drug - Adverse event}). See code available at \url{https://github.com/idiap/gme-sampler}.

%  In datasets with imbalanced entity coverage (e.g. organisms and chemicals in LOTUS), ensuring diversity in the extracted sample is a sensible objective. Then, b

%  \begin{itemize}
%      \item Need to emphasize that this method would be more extensively explored in a dedicated article.
%  \end{itemize}

% The objective is to extract a sample $S$ of $l$ documents, from an initial set $D$ of size $L$, with an optimized diversity of mentioned organisms and chemicals.

% 
\subsection{Different strategies for Relation Extraction}
% This section describes the various approaches investigated for the purpose of identifying relationships between organisms and isolated NPs from the abstracts of scientific articles.

\subsubsection{Few-shot learning with open LLM}
\label{meth:icl}
In few-shot settings, the model is prompted with $K$ input-completion example pairs and one final input, with the objective of accurately generating the completion for the final input \cite{brown_language_nodate}. Considering the limited size of the context-window (2048 tokens), we carefully selected archetypical parts of diverse abstracts that exemplify various patterns and specificities of reporting NP relationships. More details in supplementary \ref{sec:impl-details-icl}.

\subsubsection{Fine-tuning}
\label{meth:ft}
The task was framed as a special case of end-to-end relation extraction with a single relation. Several factors influenced this design: the need for a generalised NER component that is not dictionary-based, as new species and NPs are discovered and named continuously, the scarcity of training data to segment the task in a NER-RE pipeline, and, the specific decoding requirements related to the NP relationships of "Multiples" entities. Given an input text $X$ reporting relations $\{r_1, r_2, r_3\}$ between organisms $o_1$ and natural-products $c_{1:3}$ like: "Three new metabolites, gloeophyllins A-C (1-3) have been isolated from the solid cultures of Gloeophyllum abietinum.", the expected output is the linearised list of relations $Y$: "Gloeophyllum abietinum produces gloeophyllin A; Gloeophyllum abietinum produces gloeophyllin B; Gloeophyllum abietinum produces gloeophyllin C", following previous works recommendations \cite{luo_biogpt_2022, giorgi_sequence--sequence_2022}. During fine-tuning, the objective function is then defined as:
\begin{equation}
\label{loss}
    L(\theta) = \sum\limits_{t=1}^{|Y|}{\log p(y_t | X, y_{<t}, \theta)}
\end{equation}
For efficient fine-tuning with minimal memory and parameter requirements, we adopted the QLoRA approach \cite{dettmers_qlora_2023}. The method extends the Low-rank Adaptation (LoRA) technique \cite{hu_lora_2021}, which involves freezing the original model weights and training only a small set of parameters, known as adapters. Given the original weight of a layer $W$, with $h=Wx$, the adapters operate as an update to the initial weights $W + \Delta W$ and give $h = Wx + \Delta Wx$. With QLoRA, the LoRA strategy is integrated with quantization to minimize the memory footprint. It uses a low-precision storage data type ($\text{NF4}$) and a computation data type ($\text{BFloat16}$) for the forward and backward passes. Two models were evaluated using this strategy: BioGPT \cite{luo_biogpt_2022} and GPT-2 \textit{Medium} \cite{radford_language_nodate}. Both models share the same architecture, but BioGPT and its tokenizer were trained on PubMed items and achieved SOTA performance on three end-to-end RE tasks. As \citet{luo_biogpt_2022}, we also add Seq2rel \cite{giorgi_sequence--sequence_2022} as a second baseline using a sequence-to-sequence approach. See implementation details and hyperparameter tuning in supplementary \ref{sec:impl-details-ft}.

% ://github.com/ggerganov/llama.cpp#llamacpp

% The adapters modify the weights $W$ of a layer by adding the update term $\Delta W$, such that $h = WX + \Delta WX$ QLoRA extends the Low-rank Adapter (LoRA) fine-tuning method [Hu].
\subsection{Synthetic abstracts generation}
\label{sec:meth-synth-generation}
\begin{figure*}
\begin{center}
  \includegraphics[width=.9\textwidth]{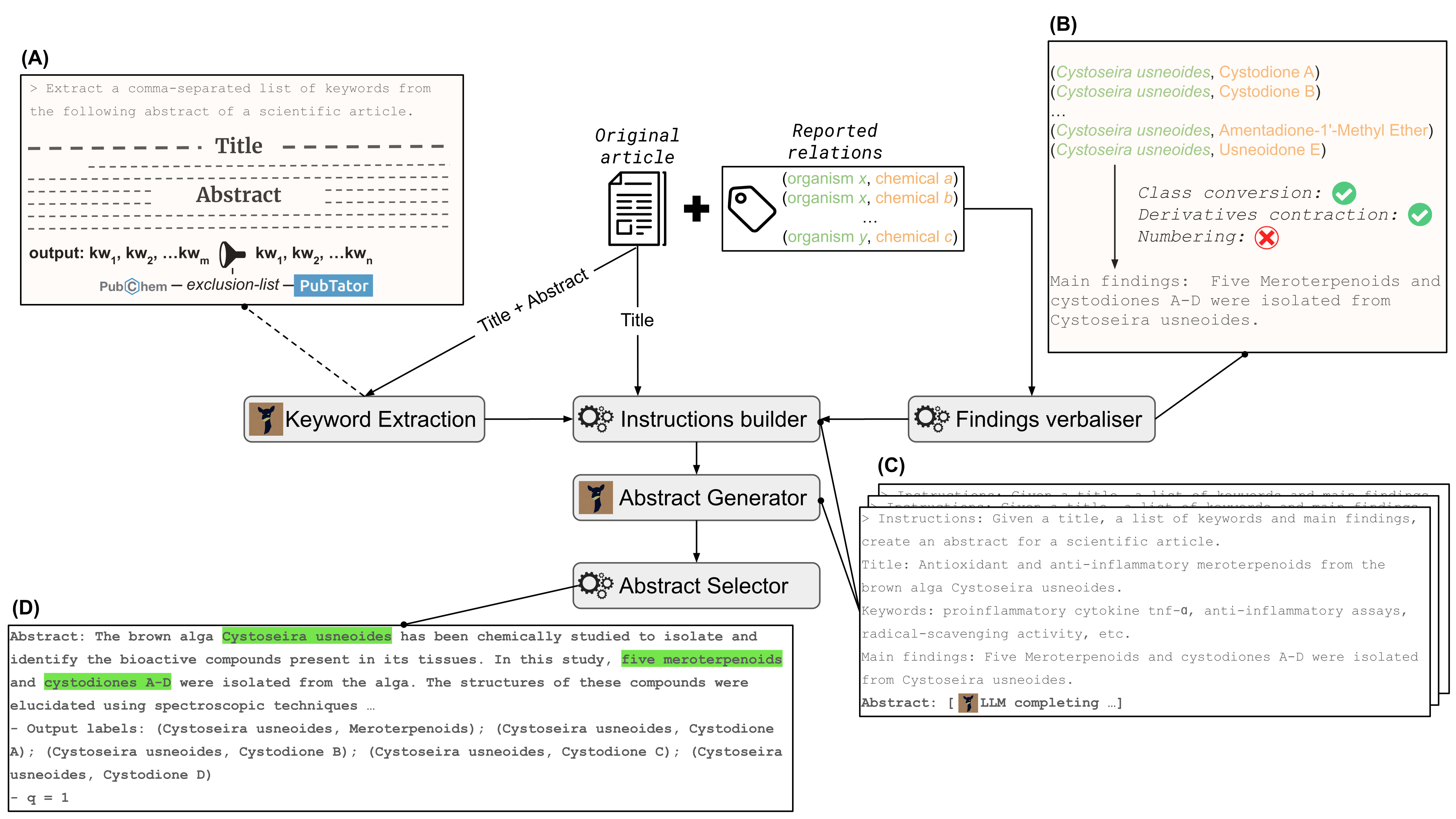}
\end{center}
\caption{Description of the synthetic abstract generation process}
\label{fig:syntheticAbstractGeneration}
\end{figure*}

A general overview of the synthetic abstract generation is provided in Figure \ref{fig:syntheticAbstractGeneration}. The goal is to leverage the generative capabilities of instructions and conversations tuned models to correct the discrepancies between the output labels and the input text. Consistency is maintained by grounding the generated abstracts on key elements from an original seed abstract: title, keyphrases extracted from the abstract (and title), and verbalized main findings. The main findings represent the set of relations $\{r_1, r_2, ..., r_n\}$ between organisms and NPs reported in the seed article. Both the keyphrase extraction and the subsequent generation step can be framed as instructions-guided tasks: "\textit{Extract a list of keywords ...}", "\textit{Create a scientific abstract ...}". As the extracted keywords and keyphrases will provide an essential context to constrain the generation of the synthetic abstracts, it is also arguably advantageous that the both tasks are carried by the same model.\\
The extraction of keywords (illustrated in Box \textbf{A} of Figure \ref{fig:syntheticAbstractGeneration}) involves prompting the model with appropriate instructions and filtering any outputs corresponding to organism or chemical names using an exclusion list. The exclusion list is built from the available PubTator's annotations \cite{wei_pubtator_2019} of named entities (if any) and the list of synonyms available from the LOTUS annotations. This filter is essential to ensure the consistency of the generated abstract by avoiding any contextual information that could interfere with the provided main findings. Together with the original title, the keyphrases provide a biological context in which to express the NP relationships.\\
By explicitly formalizing the expected patterns in upstream instructions, the expression of NP relationships during the generation step can be more efficiently controlled. The findings-verbaliser module operates as a sampler to emulate and combine various patterns of expression that can be observed in the literature. It incorporates 5 possible transformations: (1) members of a same chemical class\footnote{according to NP-classifier \cite{kim_npclassifier_2021} annotations in LOTUS} can be replaced by the simple mention of the class (e.g: a list of chemicals $c_{1:5}$ is replaced by the more concise mention "\textit{Five Meroterpenoids}"); (2) Lists of chemical derivates can be contracted (e.g "\textit{Cystodione A-D}"); (3) The order of relationships is systematically shuffled; (4) Chemicals can be numbered (e.g "\textit{Cystodione A-D} (1-4)"); (5) Directions of the relationship can change from "\textit{$O$ produces $C$}" to "\textit{$C$ was isolated from $O$}". See the box \textbf{B} of Figure \ref{fig:syntheticAbstractGeneration} and more details in supplementary \ref{sec:impl-details-generation}. These different transformations are reminiscent of the strategies commonly used in data augmentation \cite{feng_survey_2021}.\\
For each input seed abstract, $m$ instructions are sampled and assembled following this procedure and forwarded to the model for generation (Box \textbf{C}). Finally, the selector module selects a top $k$, from the $m$ generated abstracts, ensuring that at least a proportion $q$ of the expected relations have the labels of the involved organisms and chemicals explicitly mentioned in the generated abstract (Box \textbf{D}). Regarding the expected output labels, the replacement operated by transformation (1) also applies: the initial relations $r_{1:5}$ involving the 5 meroterpenoids are replaced by a single relation $r_6$ involving "Meroterpenoids" as chemical entity. In contrast, transformation (2) does not affect the output labels, requiring the model to expand the list of relations involving each derivative (see Box \textbf{D} - \textit{Output labels}). Also, the loss in eq.\ref{loss} (like in Seq2rel) is permutation sensitive, but the order created by the transformation (3), which also applied to the output labels, is almost systematically respected by the model in the generated abstract, alleviating this issue. Transformations (4) and (5) have no influence on the output labels.

%  This filter is essential to ensure the consistency of the generated abstract by avoiding any contextual information that could interfere with the provided main findings.Together with the original title, the keyphrases describe a biological context in which to express the main findings.

%------------------------------
\section{Empirical experiments}
\begin{figure*}
\begin{center}
  \includegraphics[width=.9\textwidth]{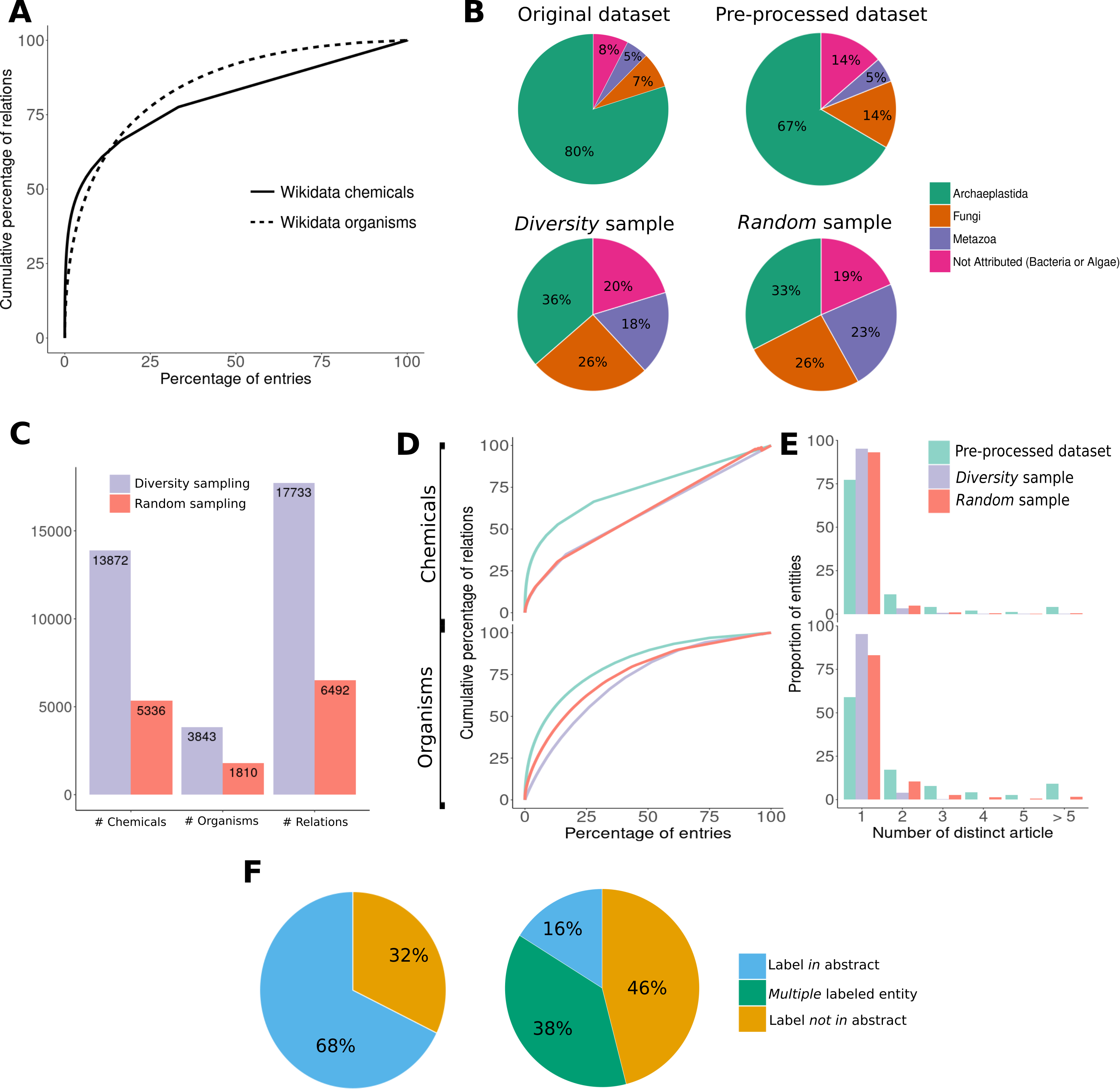}
\end{center}
\caption{\footnotesize \textbf{A:} Distribution of the cumulative proportion of reported relations per fraction of chemicals (plain line) and organisms (dashed line), ordered by their contributions. The same relations but reported from different articles are considered as distinct items. From the both curves, it can be estimated that 20\% of the most represented chemicals holds more than 68\% of the relations, and 20\% of the organisms hold more than 72\%. \textbf{B:} Repartition of the number of reported relations, organised per kingdom of the subject organism in 4 datasets: the original (\textit{full}) dataset, the pre-processed dataset, the \textit{Diversity} sample and \textit{Random} samples. For \textit{Random} samples, proportions are averaged over 5 samples. \textbf{C:} Statistics of the number of distinct organisms, relations and chemicals in the \textit{Diversity} sample compared to \textit{Random} samples. All samples contain 2000 articles. \textbf{D:} Similarly to \textbf{A}, Distribution of the cumulative proportion of reported relations per fraction of chemicals (top) and organisms (bottom), ordered by their contributions in three different type of samples: \textit{Pre-processed}, \textit{Diversity} and \textit{Random}. \textbf{E:} Distribution of the frequency of mention in distinct articles of chemicals (top) and organisms (bottom). \textbf{F:} Mismatches between standardized labels of organisms and chemicals and their original literal mentions in the abstracts of articles reporting the relationships. \textit{Multiple} entities correspond to chemical mentions that are not expressed in a continuous string. See details in supplementary \ref{sec:label-mismatches}.}
\label{datasetStats}
\end{figure*}

\subsection{Imbalanced repartition of reported relations and coverage on biological kingdoms}

As reported in the original release of the LOTUS dataset \cite{rutz_lotus_2022}, the imbalance in the data distribution manifests at two main levels: the repartition of the number of reported relationships per organism (resp. chemicals) and the coverage of biological kingdoms. These observations were reproduced from the latest available snapshot of the LOTUS dataset (v10-01-2023)\footnote{\url{https://zenodo.org/record/7534071}}, containing more than $533,000$ distinct relations between organisms and NPs, reported from more than $88,000$ articles. As expected, a small fraction of the organisms (resp. chemicals) attracts a large proportion of the relations: more than 72\% of relations involve only 20\% of the organisms (Figure \ref{datasetStats}.A). Beside these pareto distributions \cite{newman_power_2005}, the imbalance in the repartition of the relations across biological kingdoms is also important: 80\% are related to \textit{Archaeplatida} (Figure \ref{datasetStats}.B \textit{top-left}). Considering these both biases is essential to extract a valuable sample. This motivated the use of the GME-sampler in a stratified way, to maximise diversity and reduce the pareto effect, while ensuring a more balanced coverage across biological kingdoms.

\subsection{Dataset pre-processing}

The original dataset was first preprocessed and filtered prior to sampling to eliminate various sources of perturbations and unusable data in subsequent steps. Specifically, only documents with publicly available abstracts on PubMed were selected, and these were further filtered based on the number of reported relations. Indeed, a manual inspection of a subset of articles revealed that documents reporting large numbers of relations (\cite{swainston_recon_2016, thiele_community-driven_2013, stefanini_core_2017, thompson_phytoestrogen_2006}) often propose genome-scale metabolic reconstructions, large screening analyses, or database releases. Although these documents may report hundreds of relationships, they are typically not expressed in the abstracts, making them useless examples for building a RE model. Only articles reporting less than 20 relations (corresponding to the quantile 93\%) were then selected. Compared to organism names, the length of chemical names can exhibit extreme variability and exceed hundreds of characters depending on the nomenclature. To mitigate the issues posed by these lengthy labels, which are inordinate to decode and could consume an excessive portion of the context window, only relations involving chemicals with a label length $l \le 60$  characters were retained. More details in supplementary \ref{chemical-label-TH}. See the global pre-processing statistics in supplementary table \ref{tab:impactPreProcessing} and kingdom coverage in Figure \ref{datasetStats}.B top-right.

\subsection{Building a diversity-augmented dataset}

\begin{figure*}
\begin{center}
  \includegraphics[width=1.1\textwidth]{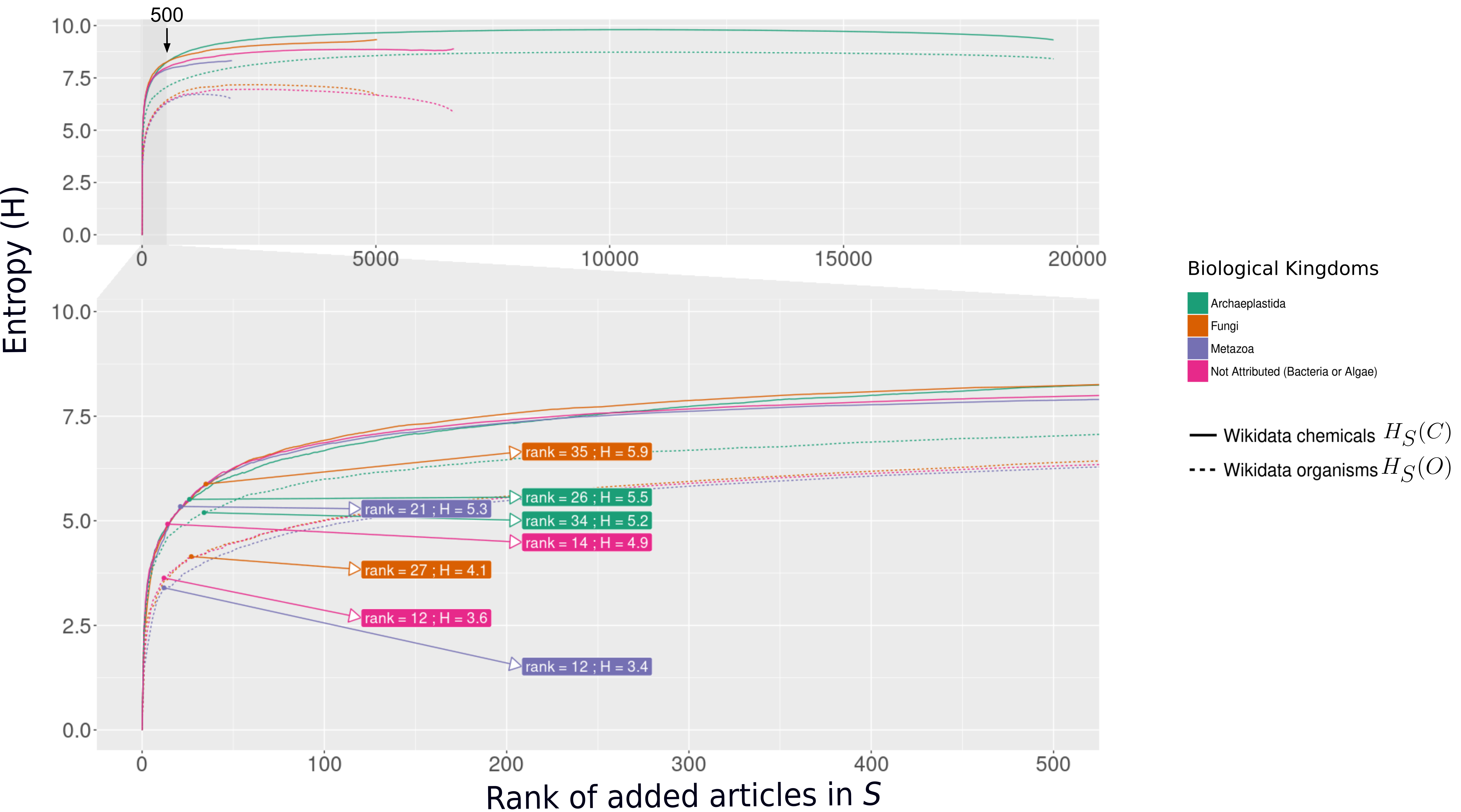}
\end{center}
\caption{\textbf{Top-panel:} Evolution of the Entropy over the distribution of reported organisms ($H_{\displaystyle S}(O)$) and chemicals ($H_{\displaystyle S}(C)$) by adding iteratively a new article ($d*$) in the built dataset, stratified by biological kingdoms. The step (500) when 80\% of the maximal entropy is reached in all the kingdom's subsets, for organisms and chemicals, is indicated with the black arrow (more details in supplementary table \ref{tab:PercentageMaxEntropy}). \textbf{Bottom-panel: } Zoom of the evolution of $H_{\displaystyle S}(O)$ and $H_{\displaystyle S}(C)$ in the first 500 added articles. For each curve, the knee-points (\textit{bending} points) with the corresponding rank and associated entropies are indicated.}
\label{entropydegradation}
\end{figure*}

\subsubsection{Diversity-sampling on organisms and chemicals}
The preprocessed dataset was first stratified according to the taxonomic classification (kingdoms) of the organisms associated with the relations reported in each document. Subsequently, the GME-sampler was applied to each subset (Figure \ref{entropydegradation}:Top-panel) to monitor the evolution of the diversity metrics ($H_S(O)$ and $H_S(C)$) and determine an optimal sample size. Indeed, the GME-sampler operates as a ranking method, where the article selected at step $n$, is the one which contributes the most to the diversity of the set of the $n-1$ articles selected upstream. For both organisms and chemicals, diversity increases rapidly in the first hundred ranked items, followed by a plateau. Specifically for organisms (regardless of the kingdom), diversity showed a decline in the second half of the sampled items (supplementary table \ref{tab:Tab-max-entropies}). This is the signal that the addition of new articles provide relations for already well-covered organisms and disrupted the existing balance in the organism distribution. In contrast, the impact of newly added articles on chemicals is negligible, likely because they represent a larger set of distinct entities (supplementary table \ref{tab:impactPreProcessing}). To keep a reasonable balance between diversity and sample size, we decided to only retain the top $n=500$ ranked articles per kingdoms, ensuring at least $80\%$ of the maximal observed entropy on both organisms and chemicals (Figure \ref{entropydegradation}:Bottom-panel). The proportions of maximal observed entropy at alternative sample sizes are presented in supplementary table \ref{tab:PercentageMaxEntropy}.\\
The impact of the diversity-sampling strategy is evaluated by comparing the composition of the sample against 5 random samples of equivalent sizes\footnote{Each random sample is composed of 500 random literature items sampled per kingdoms}. The original diversity sample and the extracted random samples are respectively denoted as \textit{Diversity} and \textit{Random} samples. While showing similar kingdoms' coverage because of the common stratification procedure (Figure \ref{datasetStats}.B bottom), the diversity sample is, as expected, significantly richer in terms of distinct number of chemicals, organisms and relations (Figure \ref{datasetStats}.C. This improved diversity is also reflected in a reduced pareto effect for the distribution of the organisms (negligible for chemicals), and overlap between the entities reported in each article (Figures \ref{datasetStats}.D and E). A more comprehensive comparison against others possible sampling strategies for diversity are discussed in supplementary \ref{sec:div-alt-strats}.

\subsubsection{Distance between standardised annotations and original text}
\label{sec:Mismatches-annot-text}
Several works emphasize the importance of data quality over quantity for fine-tuning language models \cite{zhou_lima_2023, dettmers_qlora_2023, li_quantity_2023}. LOTUS data are recognized as being of high-quality, particularly because of the harmonization, cleaning and validation steps of the workflow, aligning original records from several open NP databases into standardized structures and organisms in Wikidata. Although this is essential to ensure data FAIRness, these processes logically distance the standardised entries from their original literal mentions in the referenced articles. To get a rough estimate of this distance, the \textit{Diversity} and the 5 \textit{Random} samples were merged into a single \textit{Extended} dataset. Then, we estimated the proportion of the labels of the standardized entities that could be found in the original abstracts of articles reporting the relationships. Details of this estimation in supplementary \ref{sec:label-mismatches}. More than 2/3 of the organisms labels are effectively retrieved in the original abstract, while less than half of the chemical names can be retrieved, even considering their synonyms (Figures \ref{datasetStats}.F). Assuming that these two types of mismatches are independent, only 1/3 of the reported pairs would be completely found in an abstract. Finally, some reported NP relationships are simply not expressed in the abstract of the cited reference, but have been reported from the body of the article or supplementary materials\footnote{However, we have chosen to focus only on abstracts and not on full texts because of their much greater availability and their synthetic forms.}. Whether they are derived from the \textit{Diversity} or a \textit{Random} sample, these noisy examples make the training of a model challenging because some labels to be predicted are missing from the input text \cite{northcutt_confident_2022, jain_overview_2020}. In this context, alternative strategies like Zero-shot or Few-shot learning (also called In-Context Learning) based on open LLMs \cite{liu_what_2021, chen_meta-learning_2022} also need to be considered.

\subsubsection{Creating a manually curated evaluation dataset}
If these discrepancies certainly affect the training of a model, they are a more sensitive issue in an evaluation set \cite{northcutt_pervasive_2021}. Also, if diversity can be an important feature for a training set \cite{yu_can_nodate}, it is arguably also important for an evaluation set \cite{liang_advances_2022}. While smaller by design, the evaluation set needs to be representative. Finally, since the manual curation of an evaluation set is an expensive and time-consuming task, the selected set of entries need to be chosen carefully \cite{sambasivan_everyone_2021}. Considering the last points, the knee-points of the entropy curves (where the entropy increases weaker by new added articles) obtained with the GME-sampler suggest relevant tradeoffs between sample size and diversity (\ref{entropydegradation}:Bottom-panel) early in the sampling. Nonetheless, as they vary between different biological kingdoms, on organisms and chemicals, and could be too restrictive, the extended set of the top 50 items from each kingdom was extracted, resulting in an evaluation dataset of 200 abstracts. The abstracts were manually curated by an expert, annotating all instances of mentioned organism-NP relationships in their order of appearance in the original text and using established identifiers such as Wikidata IDs and PubChem IDs. As isolated chemicals are sometimes grouped into chemical families for the sake of brevity in abstracts, all mentions of a more general chemical family were also annotated. The curated evaluation set is publicly available at \url{https://zenodo.org/records/8422007}. Details about the curation protocol are available in supplementary \ref{curation-protocol}. The benefit of the diversity-sampling for building the evaluation set is discussed in more details in supplementary \ref{sec:curated-set-eval}.

\subsection{Few-shot learning approaches the performance of standard fine-tuning on raw data}

The mismatches between the standardized labels and the original abstracts have therefore been corrected for the evaluation set. However, due to the considerable investment of time and resources required for this task, the same corrections were not applied on the remaining data available for training. In this particular context of noisy data for end-to-end RE, two strategies were evaluated: standard fine-tuning and few-shot learning, the latter being able to rely only on a few manually selected examples. The performance of the fine-tuning strategy were evaluated based on all available examples (\texttt{Full}) and 3 alternatives train/valid datasets derived from the initial \textit{Diversity}, \textit{Random} and \textit{Extended} samples, here after referred as: \texttt{Diversity-raw}, \texttt{Random-raw} and their union \texttt{Extended-raw}. Their respective sizes and splits are detailed in supplementary table \ref{tab:statsDatasets}. All datasets were used to train 3 models for end-to-end RE: Seq2rel, BioGPT and GPT-2. Six open LLMs were also evaluated in few-shot learning settings: LLaMA 7B, 13B, 30B and 65B, along with two models, respectively fine-tuned on instructions and conversations and derived from LLaMA 7B and 13B: Alpaca-7B and Vicuna-13B.\\
Best performance in fine-tuning settings was achieved by BioGPT (Table \ref{tab:FTvsICL}). Regardless of the training dataset\footnote{Except for \texttt{Full}}, it consistently outperformed Seq2rel and GPT-2 and demonstrated an $\text{f1-score}$ of $32.5\%$ when trained on \texttt{Extended-raw}. We also evaluated the influence of the different training datasets on models performance. The results indicate that models trained on \texttt{Diversity-raw} outperformed those models trained on \texttt{Random-raw}, with a notable improvement in recall at the expense of precision. Merging the datasets into a larger (\texttt{Extended-raw}) also resulted in improved performance for all models. However, expanding the dataset to all available examples only barely improved the previous performance and surprisingly underperformed with BioGPT. In few-shot learning scenarios, the best performance was obtained with LLaMA-65B and decline with smaller models. Although the performance was inferior compared to fine-tuned alternatives, the models achieved reasonable scores considering the limited number of archetypal examples provided. These results also emphasize the potential of few-shot learning or prompt-tuning based approaches in practical context with low-resources.

% \begin{itemize}
%     \item Compare performances of a seq2rel and BioGPT-Qlora fine-tuned model on the \textit{Extended-raw} dataset versus in 5-shot in-context learning on different LLM.
%     \item ICL slightly less better, but only considered 5 archetypal sentences as "\textit{examples}"
% \end{itemize}

\begin{table}[ht]\centering
\scriptsize
% I had to add & at the end of each rows the have the lateral bar
\begin{tabular}{lrrrrr}\toprule
\textbf{model} &\textbf{Training} &\textbf{precision} &\textbf{recall} &\textbf{f1} &\\
\midrule
\textbf{LLaMA-7B} &\multirow{6}{*}{Few-shot learning (5-shot)} &27.0 &9.04 &13.55 &\\
\textbf{LLaMA-13B} & &35.64 &23.64 &28.49 &\\
\textbf{LLaMA-30B} & &38.51 &\ul{23.24} &28.99 &\\
\textbf{LLaMA-65B} & &\ul{40.16} &22.97 &\ul{29.23} &\\
\textbf{Alpaca-7B} & &15.14 &2.21 &5.86 &\\
\textbf{Vicuna-13B} & &38.4 &20.43 &26.48 &\\
\midrule
\multirow{3}{*}{\textbf{Seq2rel}} &\texttt{Random-raw} &43.2 +/- (6.67) &4.8 +/- (1.16) &8.6 +/- (2.00) &\\
&\texttt{Diversty-raw} &39.6 &5.4 &9.5  &\\
&\texttt{Extended-raw} &\textbf{47.3} &5.8 &10.4 &\\
&\texttt{Full} &45.6 &7.1 &12.2 &\\
\midrule
\multirow{3}{*}{\textbf{GPT-2}} &\texttt{Random-raw} &32.5 +/- (4.83) &11.8 +/- (5.25) &15.0 +/- (2.54) &\\
&\texttt{Diversty-raw} &22.3 &19.2 &20.6 &\\
&\texttt{Extended-raw} &44.8 &21.7 &29.3 &\\
&\texttt{Full} &47.5 &22.5 &30.5 &\\
\midrule
\multirow{3}{*}{\textbf{BioGPT}} &\texttt{Random-raw} &47.2 +/- (4.01) &19.8 +/- (2.71) &27.6 +/- (2.48) &\\
&\texttt{Diversty-raw} &37.1 &\textbf{28.4} &32.2 &\\
&\texttt{Extended-raw} &42.2 &26.5 &\textbf{32.5} &\\
&\texttt{Full} &46.7 &21.3 &29.3 \\
\bottomrule
\end{tabular}
\caption{Performance of 5-shot \textit{in-context} learning using LLaMA and LLaMA-instructed models compared to \textit{seq2rel}, GPT-2, BioGPT finetuned models. Three types of training dataset are evaluated: the diversity sample (\texttt{Diversty-raw}), 5 random samples (\texttt{Random-raw}) and the extended sample (\texttt{Extended-raw}), which is the union of the previous samples. \texttt{Full} is a dataset contained all available examples from the LOTUS snapshot, except the 200 used in the evaluation set. Best performance via finetuning are bold, while best performance in few-shot settings are underlined.}
\label{tab:FTvsICL}
\end{table}
\subsection{Reversing the task: Generation of synthetic data with open LLMs}

While LLMs cannot compete in terms of performance with fine-tuned approaches in the evaluated settings, their generative abilities could be used alternatively to address the main bottleneck: the discrepancies between the input text and the labels in the training data. It requires going beyond distant supervision or data augmentation \cite{feng_survey_2021, shang_learning_2018, smirnova_relation_2018}. The former involves mapping relationships from a Knowledge Base to a large corpus of text to generate pseudo-labels, whereas the latter entails applying a range of transformations, permutations, or morphings to a core set of high-quality examples. In contrast, the adaptive described approach propose to generate a set of synthetic input abstracts from a pre-defined context and a set of expected output labels (i.e organism - NP relationships).\\
To maintain consistency, each synthetic abstract is based on the context and results reported from an original seed abstract. The first step is to generate the instructions to prompt the selected LLM for generation. The instructions are composed of a title, a list of keywords and the verbalised main findings (Method \ref{sec:meth-synth-generation}). We decided to use the open source Vicuna-13B \cite{chiang_vicuna_2023}\footnote{version v1.3 from 22/06/2023: \url{https://huggingface.co/lmsys/vicuna-13b-v1.3}}, a LLaMA-13B model fine-tuned on user-shared conversations collected from ShareGPT\footnote{\url{https://sharegpt.com/}}, which outperforms alternatives of equivalent sizes on several benchmarks \cite{dettmers_qlora_2023}. For each input seed abstract, the top-10 extracted keywords were used in the built instruction. As this is a crucial step, the performance of Vicuna-13B to extract keywords have been evaluated on the SemEval2017-Task10 dataset \cite{augenstein_semeval_2017} in supplementary \ref{sec:kw-extraction}. To diversify the generated abstracts, $m=10$ instructions prompts with different verbalisation patterns were then sampled per initial seed article. Finally, only the top $k=3$ most relevant synthesized abstracts per seed were selected with the simple, yet effective, selector module.\\
To evaluate the impact of diversity-sampling on the seed articles used for synthetic generation, we created two new datasets: \texttt{Diversity-synth} and \texttt{Random-synth}, derived from the original abstracts in the \texttt{Diversity-raw} and \texttt{Random-raw} datasets, respectively. Several illustrative examples of synthetic abstracts from \texttt{Diversity-synth} are discussed in supplementary \ref{sec:generation-examples}, highlighting both the variability and the potential caveats (errors, hallucinations) of the process. As with the original data, \texttt{Diversity-synt} and \texttt{Random-synt} were merged in \texttt{Extended-synt} to measure the impact of the dataset size. Statistics of the generated datasets are presented in supplementary table \ref{tab:statsDatasets}. In total, more than 25,000 synthetic abstracts were generated from the 7901 originally contained in the \texttt{raw} datasets. From \texttt{Diversity-raw}, 200 initial items were excluded by the selector module and 162 on average for \texttt{Random-raw}. While the distinct numbers of entities/relations dropped in synthetic datasets, the selector guarantees that these labels are part of the generated abstracts. Furthermore, the generation process enables the integration of examples with chemical classes in the input text and expected labels, which were not available in the original data.\\

% Pour moi la principale différence avec la data augmentation (DA) c'est qu'en DA classique on assume que l'on a un set d'eexemple propore depuis lesquels dérvés de nouveaux ici non 

% \begin{itemize}
%     \item Show some examples of creation prompt and outputs, with the related caveats (TODO: trouver genre 3 exemples intéressant et les mettre en sups.)
%     \item Refer to the KW-step evaluation on SemEval2017 (see \ref{tab:evalKWExtraction}).
%     \item Discuss the statistics on the generated datasets. (Some excluded articles, etc.) and estimated the total number of "kept" generation. Just on average, no need of a table, I think. (Cf. \ref{tab:statsSyntheticDataset})
% \end{itemize}

\subsection{Training on synthetic data improved performance over noisy raw data}

The synthetic datasets were used to train new instances of the previously evaluated models: Seq2rel, GPT-2 and BioGPT. Although the synthetic training sets (\texttt{Diversity-synt} and \texttt{Random-synt}) are almost half the size of \texttt{Extended-raw} (resp. 3562 and 3798 compared to 7111 examples), on which was established the previous baseline with BioGPT ($\text{f1-score}=32.5$), all the trained models demonstrated improved performance (see Table \ref{tab:FTSyntheticData}). The ranking of the models and the impact of the synthetic training sets on the final performance align with the previous observations on the original data. BioGPT models persistently outperformed Seq2rel and GPT-2, and the training on \texttt{Diversity-synt} resulted in an improved recall at the expense of precision compared to \texttt{Random-synt}. However, the GPT-2 models trained on \texttt{Random-synt} on average outperformed the one trained on \texttt{Diversity-synt}, a departure from the trend observed with Seq2rel and BioGPT. Again, the best performance is achieved by BioGPT trained on the merged set, with $\text{f1-score}=53.8$.\\
Finally, two BioGPT-Large models were trained on the \texttt{Diversity-synt} and \texttt{Extended-synt} (see Table \ref{tab:BioGPT-Large-eval}). The model trained on \texttt{Diversity-synt} achieved $\text{f1-score}=57.2$, comparable to the new best model trained on the much larger merged set ($\text{f1-score}==59.0$) and also demonstrated a better recall ($56.90$ against $51.6$).

% \begin{itemize}
%     \item 
%     \item Evaluation of the model trained on synthetic data.
%     \item While the synthetic datasets are smaller (3562 and 3798 examples for \textit{diversity} and \textit{rdm} against 7111 in the \textit{Extended-raw}) all tested approaches (Seq2rel and BioGPT-QLoRa) perform better than their equivalent on the original noisy data.
%     \item The proposed BioGPT-QLoRa model performs better than Seq2rel
%     \item Diversity datasets tends to increase recall at the price of precision.
%     \item To try to achieve the best performances, the 5 random synthetic datasets + the synthetic diversity dataset were merged and used to train a BioGPT AND a BioGPT-LargeQLoRa -> Discuss the performances of the best compared to finetuning on noisy data or ICL. 
    
% \end{itemize}

\begin{table}[H]\centering
\scriptsize
\begin{tabular}{lrrrrr}\toprule
\textbf{model} &\textbf{Dataset} &\textbf{precision} &\textbf{recall} &\textbf{f1} &\\
\midrule
\multirow{3}{*}{\textbf{Seq2rel}} &\texttt{Random-synt} &\ul{62.4 +/- (1.03)} &26.8 +/- (1.96) &37.5 +/- (1.90) &\\
&\texttt{Diversty-synt} &61.5 &30.7 &40.1 &\\
&\texttt{Extended-synt} &\textbf{65.1} &29.9 &41.0 &\\
\multirow{3}{*}{\textbf{GPT-2}} &\texttt{Random-synt} & 42.6 +/- (2.89) &32.7 +/- (2.81) &37.2 +/- (2.80) &\\
&\texttt{Diversty-synt} &28.5 &39.4 &33.0 &\\
&\texttt{Extended-synt} &52.0 &44.6 &48.0 &\\
\multirow{3}{*}{\textbf{BioGPT}} &\texttt{Random-synt} &56.4 +/- (2.26) &38.8 +/- (1.92) &46.0 +/- 1.08 &\\
&\texttt{Diversty-synt} &52.5 &\ul{41.2} &\ul{46.2} &\\
&\texttt{Extended-synt} &63.7 &\textbf{46.5} &\textbf{53.8} &\\
\bottomrule
\end{tabular}
\caption{Performance of Seq2rel, GPT-2 and BioGPT models fine-tuned on synthetic data. Three types of training dataset are evaluated: the diversity sample (\texttt{Diversty-synt}), 5 random samples (\texttt{Random-synt}) and the extended sample (\texttt{Extended-synt}), which is the union of the previous samples, all synthetically generated from the corresponding seed original samples. For \textit{Random-synt} samples, results are averaged and standard deviations are reported. Best performance are bold, and second best performance are underlined.}
\label{tab:FTSyntheticData}
\end{table}

\begin{table}[H]\centering
\scriptsize
\begin{tabular}{lrrrrr}\toprule
\textbf{model} &\textbf{Dataset} &\textbf{precision} &\textbf{recall} &\textbf{f1} &\\
\midrule
BioGPT-Large &\texttt{Diversty-synt} &57.50 &\textbf{56.90} &57.20 &\\
BioGPT-Large &\texttt{Extended-synt} &\textbf{69.0} &51.6 &\textbf{59.0} &\\
\bottomrule
\end{tabular}
\caption{Performance of finetuned bioGPT-Large models on the synthetic diversity sample and the Extended synthetic sample.}
\label{tab:BioGPT-Large-eval}
\end{table}

%------------------------------
\section{Discussion}
The application of deep learning models for the completion of biomedical knowledge bases is largely limited by the availability and quality of domain-specific labelled data \cite{liang_advances_2022}. Therefore, we adopted a data-centric methodology \cite{mazumder_dataperf_2023, zha_data-centric_2023}. In order to address the data imbalance and optimised the manual curation process, we proposed the GME-sampler inspired by diversity metrics commonly used in ecology. The sampler was applied on the pre-processed LOTUS dataset (separately on each biological kingdom) to extract a subset of documents, ensuring a diverse set of organisms and chemicals in the reported relations. The compositional analysis revealed a higher number of distinct entities in the extracted sample, but also a better balance considering the fixed number of items. Diversity has been recognized as an important factor in training set for representation learning and improving the generalization performance of models \cite{gong_diversity_2019, yu_can_nodate}, essential for the NER sub-task. By forcing diversity into the relation partners (organisms and chemicals), we also expect it to be improved in their mentioning contexts. Considering the time and domain expertise requirements to annotate an evaluation dataset, the diversity metric was also used for partitioning. We extracted and manually annotated a representative subset by extracting the 200 top-diverse items. We hope that this manually-curated evaluation dataset will help the community to build upon this work.\\
Despite a smaller number of trained parameters, BioGPT and GPT-2 fine-tuned with QLoRa clearly outperformed Seq2rel (see supplementary \ref{sec:impl-details-ft}). This highlighted the benefit of the larger pre-training, but also the effectiveness of the QLoRA strategy, where low-rank updates of a large, but quantized model, achieve better performance than the full fine-tuning of a smaller model, for a lower parameter budget \cite{aghajanyan_intrinsic_2020, dettmers_8-bit_2022, hu_lora_2021}. While based on the same architecture, improvements of BioGPT over GPT-2 can be attributed both to the pre-training on PubMed and also to the dedicated tokenizer (see supplementary figure \ref{fig:TokenizedAbstractDiff}). Beyond the architectures of the models, the training dataset also had a significant impact on the performance. A comparison between models trained on the largest (\texttt{Extended-raw}) and the diversity-optimized dataset revealed that the latter achieved competitive results despite its smaller size. Additionally, results also suggest that improving the diversity of the provided set of examples for training can improve the recall of the model, at the expense of precision. Increasing the number of training examples with noisy data also have limited benefits, as suggested by the comparison with the training dataset extended to all available data (no sampling, no stratification) \cite{salhofer_impact_2022, prusa_effect_2015, liang_advances_2022}. Finally, few-shot learning techniques leveraging open LLMs exhibit reasonable performance (see LLaMA-65B) and can be particularly valuable when only limited or noisy data are available. However, their larger size may incur higher management costs, necessitating careful consideration of resource allocation.\\
Instead of using them to directly perform the task, we then propose to use them to generate synthetic examples and alleviate the noise of the dataset. However, evaluating the quality of the generated abstracts is challenging. Although the process is prone to hallucinations, the factuality is not the key criteria, as long as the generated texts are credible, meaning that they are coherent and adhere to the established syntax, style and patterns, of expression of the relations in human-written abstracts. Since the training sets of LLMs contain scientific articles and abstracts, they have absorbed their stylistic and syntactic specificities. The generation of synthetic data could then be seen as a form of knowledge distillation. Moreover, while previous studies have suggested that LLMs may not be knowledgeable \cite{cao_knowledgeable_2021, si_prompting_2023, mallen_when_2023}, other investigations have highlighted the remarkable capabilities of chatbot and instructions-tuned models in following style instructions \cite{pu_chatgpt_2023, chia_instructeval_2023}. Then, a relevant evaluation criteria for these synthetic data is the improvement on the performance they provided.\\
The study revealed that the trained model exhibited a substantial increase in performance when transitioning from raw noisy data to synthetic data. This shift in data sources did not alter the previously observed trend: BioGPT outperformed other models, and the diversity-optimised sampling had a positive effect on the recall of trained models when used to select the seed articles. Most importantly, we noticed that the transition from original to synthetic data had a more determinant impact on the performance improvements than the choice of the model architecture (Seq2rel, GPT-2, BioGPT). For instance, the influence of synthetic data on the performance of BioGPT and GPT-2 is greater than the difference between the two fine-tuned models. The performance of Seq2rel was also enhanced almost by a factor of 4, notably narrowing the gap with GPT models. Similarly, scaling-up the architecture with Bio-Large ($>4.5 \times$ larger) indeed resulted in improved performance, but comparable to the previous enhancement obtained with synthetic data. Also, we noticed a clear impact of the training dataset, with the best observed recall achieved with \texttt{Diversity-synt}. These results also support the data-centric view, even in low-resource scenarios, by demonstrating improved performance over other strategies such as few-shot learning \cite{xu_towards_2022}.

% This strategy has been explored in recent studies, which demonstrate the potential of LLMs as synthetic data generators, see  \cite{abonizio_inpars_2023, jeronymo_inpars-v2_2023, tang_does_2023, josifoski_exploiting_2023, veselovsky_generating_2023}. While they focus on the general case of openIE using OpenAI LLMs, we focus more specifically on the natural-products literature using Open LLMs.
\section{Limits and future work}

Fine-tuned methods exhibit superior performance compared to zero-shot/few-shot approaches. However, the basic prompting approach used in the experiments may not fully demonstrate the capabilities of the models, and alternative strategies have been proposed \cite{zhao_calibrate_2021, wu_self-adaptive_2022, liu_what_2021}. Nevertheless, \citet{gutierrez_thinking_2022} noted that even with these improvements, the models still lack the accuracy of fine-tuned approaches with qualitative data. The use of LLMs to generate abstracts also has some evident limitations. The generated abstracts exhibit a narrow range of styles to express the relationships between organisms and chemicals compared to human-written abstracts. We suppose that the synthetic data mostly improved the recognition of organism and chemical entities, this subtask being inherently embedded in the ultimate task of decoding the relationships.\\
Although the proposed framework is effective, it cannot guarantee the true diversity of the generated abstracts and the final selection may be very similar. Secondly, the selector module does not ensure that the relations are semantically expressed in the generated abstracts, as it only checks for the explicit mention of the entities. Finally, all generated examples are designed as "positive" cases, meaning that a relation is always expected, which may not be the case in practical applications. The developed models are intended for use on a large corpus of articles and the input documents can be either selected by an upstream retriever component, or, the predictions can be re-evaluated by a downstream selector. Continuing with this data-centric view, future works will prioritize improving the three key components (instructions builder, generator, and selector) to improve the diversity of the synthetic abstracts, rather than focusing on the architecture of the trained models.\\
Given the highly dynamic nature of the LLMs research area, we anticipate significant advancements in model architecture and accessibility to arise from the research community. At the date of writing, the release of Llama (and Llama2) has paved the way for the creation of more open-license models, such as the next-generation of Vicuna\footnote{\url{https://huggingface.co/lmsys/vicuna-13b-v1.5}} or PMC-Llama trained on the biomedical literature. The development of multilingual open LLMs \cite{workshop_bloom_2023} also offers opportunities for synthetic data generation in promising areas, such as the extraction of plant-disease relationships from Traditional Chinese Medicine prescriptions, where the scarcity of labelled data is limiting \cite{li_ltm-tcm_2022}.

% To enhance the diversity of the generated abstracts, we proposed a simple strategy that involved sampling 10 instruction prompts with different verbalization patterns and then selecting the top 3 generated abstracts. 
\section{Conclusion}
With the aim of assisting the completion of NP databases, we provide the first training and evaluation datasets along with the first trained models for end-to-end RE of relationships between organisms and chemicals. Along with these main results, we explored different strategies and proposed new developments to address the problematics raised in this biomedical context. We empirically showed the benefit of the proposed GME-sampler for building a diverse and balanced evaluation dataset, as well as its positive impact on the recall via the training data. The results also indicate that the opportunities brought by the open LLMs in scenarios with few data or weakly labelled, may not lie only in their zero/few-shot learning abilities, but also in their great potential as synthetic data generator. They could open the door for the extraction of previously unexplored relationships between biomedical entities expressed in the literature, a prerequisite to unlock new paths of inferences in knowledge discovery.

%------------------------------
\section*{Funding}
This work was supported by the IDIAP Research Institute and has been done in collaboration with the company Inflamalps SA and is supported by the Ark Foundation. This project has received funding from the European Union's Horizon 2020 research and innovation programme under grant agreement No 965397. The funding bodies played no role in the design of the study, research, writing and publication of the paper. 

\section*{Acknowledgements}

The authors are thankful to Vincent Mutel, Joël Dumoulin, Joel Rossier and Colombine Verzat for their help during the project. We are grateful to Olena Hrynenko for proofreading the mathematical formulations. We are also grateful to the authors behind LOTUS, BioGPT and Seq2rel for sharing their data or code.

% \section*{References}

% \bibliographystyle{bmc-mathphys}
\bibliography{body}

\newpage

% \setcounter{page}{1}

% \setcounter{title}{1}
% \maketitle

% \appendix

\beginsupplement
% \inmput{Supplementary/SupplementaryMethods}
\section{Experimental setup and implementation details}
\label{subsec:setup-and-details}

\subsection{Few-shot learning details}
\label{sec:impl-details-icl}
\begin{figure}[H]
\begin{center}
  \includegraphics[width=1\textwidth]{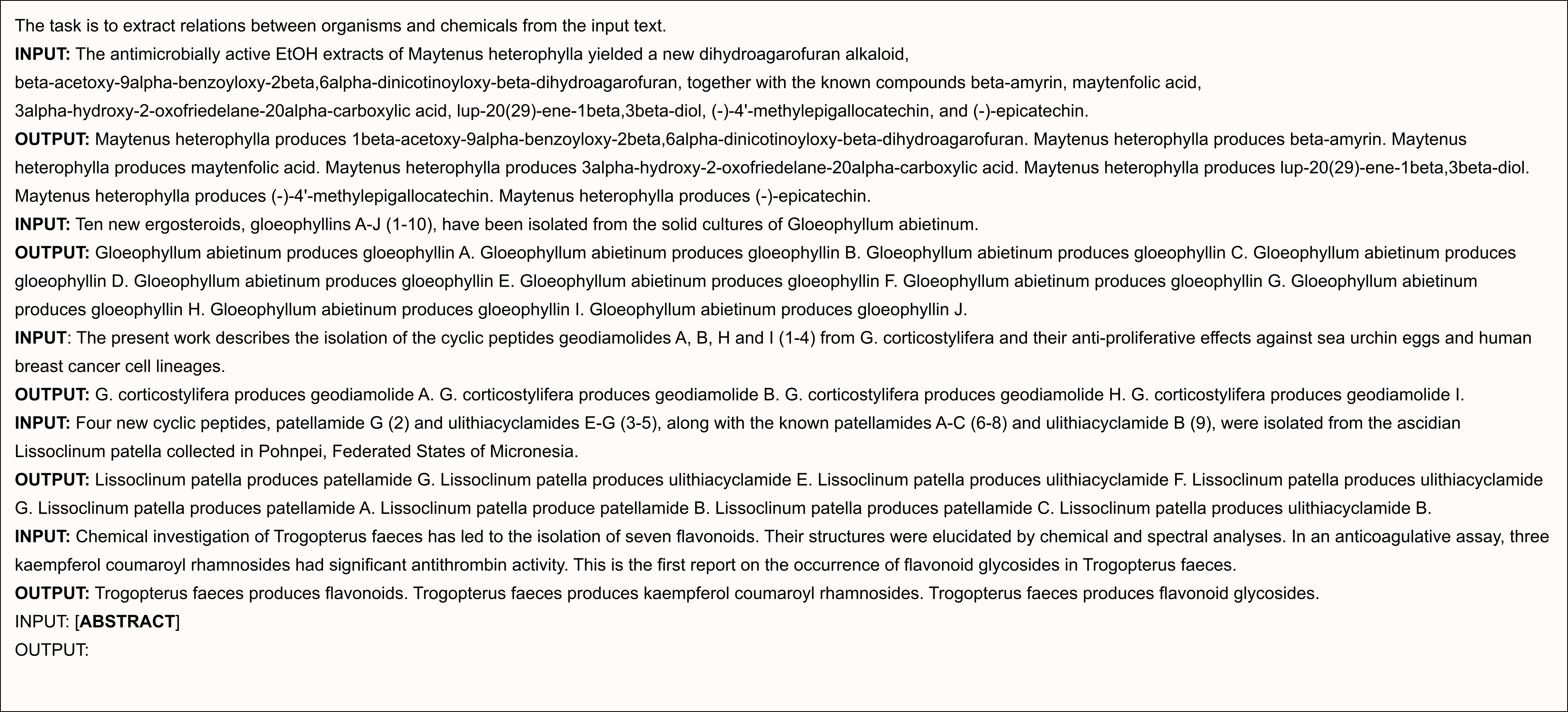}
\end{center}
\caption{Few-shot learning prompt containing five archetypical examples used on the Llama 7B, 13B, 30B and 65B.}
\label{fig:prompt-llama}
\end{figure}

Considering their particular fine-tuning, small adjustments were provided to the prompt presented in Figure \ref{fig:prompt-llama} for Alpaca-7B and Vicuna-13B. All models were also quantized for memory efficient inferences (Table \ref{tab:quantizations}). Considering our available resources, we were not able to use the \texttt{q8} (8 bits) quantization for LLaMA models $> 13B$ and improvements in performance could then be expected. In parallel, we noticed significant performance degradations when using $q4$ (4 bits). We used \texttt{llama.cpp}\footnote{\url{https://github.com/ggerganov/llama.cpp}} for quantization and inferences.

\subsection{Fine-tuning details}
\label{sec:impl-details-ft}
\citet{dettmers_qlora_2023} demonstrated the efficacy of the QLoRA approach by showing that the loss in performance due to quantization can be fully recovered through subsequent fine-tuning of the adapters and that increasing the number of adapters is crucial to match full fine-tuning performance. By exploiting the memory benefits of the \texttt{NF4} data type, we applied LoRA adapters to all linear blocks (except the initial embeddings layer) of the BioGPT and GPT-2 models. Details on the number of trained parameters are presented in Table \ref{tab:nb-trainable-params}. During training, the special tokens \texttt{<BOS>} and \texttt{<EOS>} are used to delimitate the input $X$ and the expected linearised output $Y$, such as [$X$, \texttt{<EOS>} \texttt{<BOS>}, $Y$, \texttt{<EOS>}]. The \texttt{<BOS>} token triggers the RE task at inference time.\\
For BioGPT, we evaluated different hyperparameter settings using Optuna \cite{akiba_optuna_2019}. We applied the TPE (Tree-structured Parzen Estimator) algorithm with a pruner on median to speed up the process. The hyperparameters tuning was done on the \texttt{Diversity-synt} dataset, using the $\text{f1-score}$ on the validation set as evaluation criteria. As in \citet{giorgi_sequence--sequence_2022}, we used greedy decoding during the hyperparameter tuning step and subsequently tune the decoding strategy on the best obtained configuration. The list of hyperparameters is presented in Table \ref{tab:HPTuningBioGPT}. No significant impact of the batch-size on the performances was noticed, although larger values tends to improve the stability of the training. As already observed in  \cite{aghajanyan_intrinsic_2020, hu_lora_2021}, increasing the rank of the LoRA adapters from $r=8$ to $r=16$ only led to marginal improvements. The best configuration obtained for BioGPT was reused for BioGPT-Large and GPT-2. For all evaluated datasets, models were then trained during 15 epochs (10 for BioGPT-Large) with 100 warm-up steps and the best epoch was selected using the validation set. We used the available implementation of QLoRA with PEFT \cite{peft}. We used the recommended $8-bits$ paged AdamW optimizer\footnote{\url{https://github.com/TimDettmers/bitsandbytes}} \cite{dettmers_8-bit_2022}.\\
Similarly, we tuned the hyperparameters of Seq2rel on the \texttt{Diversity-synt} dataset (See Table \ref{tab:HPTuningSeq2Rel}). All the fine-tuning experiments were conducted on an NVIDIA GeForce RTX 3090.

\subsection{Main-findings verbalisation patterns}
\label{sec:impl-details-generation}
To emulate different patterns of expression of the NP relationships, 5 transformations are applied: (1) chemical class replacement, (2) derivates contraction, (3) shuffling; (4) numbering, (5) relation directionality. The findings-verbaliser module operates as a sampler, and each transformation has an assigned probability. In the conducted experiments, we used $p_1=0.2$, $p_2=0.9$, $p_3=0.25$ and $p_4=0.9$ for the corresponding transformations. The values were empirically estimated from observed behaviours in the literature. To enhance the diversity of the generated abstracts, the temperature parameter is also randomly sampled at each generation: $t \in \{0.5, 0.6, 0.7, 0.8\}$, as similarly evaluated by \cite{chung_increasing_2023}. Similarly to few-shot learning, we also used \texttt{llama.cpp} through the python bindings library \texttt{llama-cpp-python}\footnote{\url{https://github.com/abetlen/llama-cpp-python}} for inference in generating the synthetic abstracts.

\subsection{Evaluation details}
\label{sec:impl-details-evaluation}
All evaluated models (in fine-tuning and few-shot settings) were evaluated for end-to-end RE, jointly performing NER and RE, framed as a generative task. The performances of the tested models were assessed by measuring the \texttt{f1-score} over the predicted relations extracted from the decoded outputs. An extracted relation is considered correct only if the head (an organism) and the tail (a chemical) entities exactly match the ground-truth labels.

\begin{table}[H]\centering
\scriptsize
\begin{tabular}{lrr}\toprule
\textbf{model} &\textbf{Quantization type (size in GB)} \\\midrule
\textbf{Llama-7B} &q8 (6.8 GB) \\
\textbf{Llama-13B} &q8 (13.2 GB) \\
\textbf{Llama-30B} &q5\_K\_M (21.9GB) \\
\textbf{Llama-65B} &q5\_K\_M (44.1GB) \\
\textbf{Alpaca-7B} &q8 (6.8 GB) \\
\textbf{Vicuna-13B} &q8 (13.2 GB) \\
\bottomrule
\end{tabular}
\caption{Table of quantizations procedures applied on the used LLMs. See \url{https://github.com/ggerganov/llama.cpp}.}
\label{tab:quantizations}
\end{table}

\begin{table}[H]\centering
\scriptsize
\begin{tabular}{lrrr}\toprule
&\textbf{Total parameters} &\textbf{Trainable parameters} \\\midrule
\textbf{Seq2rel} &118546185 &118546185 (100\%) \\
\textbf{BioGPT} &350649472 &3886208 (1.11\%) \\
\textbf{BioGPT-Large} &1582722536 &11533736 (0.73\%) \\
\textbf{GPT-2 \textit{Medium}} &358381208 &3555992 (0.99\%) \\
\bottomrule
\end{tabular}
\caption{Statistics of the number of trainable parameters per evaluated models.}
\label{tab:nb-trainable-params}
\end{table}

\begin{table}[H]\centering
\scriptsize
\begin{tabular}{lrrr}\toprule
&\textbf{Tuned ?} &\textbf{Value} \\\midrule
\textbf{Training} &  &  \\
\texttt{Batch size} &yes &16 (12) \\
\texttt{Number of epochs} & no &15 \\
\texttt{LoRa r} &yes &8 \\
\texttt{LoRa alpha} &yes &16 \\
\texttt{Learning rate} &yes &1.00e-4 \\
\texttt{Weight decay} &no &0.01 \\
\texttt{Gradient accumulation steps} &no* &5 \\
\texttt{LoRA dropout} &no &0.05 \\
\texttt{LoRa target modules} &no &q\_proj, k\_proj, v\_proj, out\_proj, fc1, fc2, output\_projection \\
\textbf{Decoding} &  &  \\
\texttt{strategy} &yes &beam search \\
\texttt{beam size} &yes &3 \\
\texttt{stopping criteria} &yes &never \\
\texttt{length penality} &yes &1.5 \\
\texttt{temperature} &no &0 \\
\bottomrule
\end{tabular}
\caption{Hyperparameters values used for BioGPT. Hyperparameters were fine-tuned using Optuna on the \textit{Diversity-synt} dataset. Values between parentheses correspond to adaptation for BioGPT-Large. The following settings were evaluated: \texttt{batch-size} $\in \{4, 8, 16\}$; \texttt{learning-rate} $\in [1e-6; 1e-3]$; \texttt{LoRA configurations} $\in \{(r=4, \alpha=4), (r=4, \alpha=8), (r=4, \alpha=16), (r=8, \alpha=8), (r=8, \alpha=16), (r=8, \alpha=32), (r=16, \alpha=16), (r=16, \alpha=32), (r=,16 \alpha=64)\}$. \texttt{Gradient accumulation steps} values were directly scaled in inverse proportion to the \texttt{batch-size}: $\{ 20, 10, 5\}$. For the decoding strategies, the following settings were also evaluated: \texttt{beam-size} $\in \{3, 5\}$; \texttt{stopping criteria}  $\in \{ \text{True}, \text{False}, \text{never}\}$; \texttt{length penality} $\in [0, 3]$.}
\label{tab:HPTuningBioGPT}
\end{table}

\begin{table}[H]\centering

\scriptsize
\begin{tabular}{lrrr}\toprule
&\textbf{Tuned ?} &\textbf{Value} \\\midrule
\textbf{Training} & &// \\
decodr learning rate &yes &9.00e-4 \\
batch size &no &4 \\
number of epochs &yes &20 \\
gradient accumulation steps &no &10 \\
others &no &identifical to seq2rel's CDR config \\
\textbf{Decoding} & &// \\
beam size &yes &5 \\
length penality &yes &1 \\
\bottomrule
\end{tabular}
\caption{Hyperparameters values used for Seq2rel. Hyperparameters were fine-tuned using Optuna on the \textit{Diversity-synt} dataset. All non-mentioned parameters were set according to the CDR configuration in the original Seq2rel's paper. The following configurations were evaluated: decoder's \texttt{learning-rate} $\in [1e-6; 1e-3]$; \texttt{beam-size} $\in [3,5]$; \texttt{length penality} $\in [1, 3]$.}
\label{tab:HPTuningSeq2Rel}
\end{table}

\section{Supplementary materials}

\subsection{Chemical lengths thresholding}
\label{chemical-label-TH}
To determine a reasonable threshold for filtering chemical labels with excessive length, we conducted a comparative analysis of the distribution of label lengths in LOTUS (derived from Wikidata) versus their corresponding IUPAC names (See supplementary Figure \ref{chemicalnchar}). While the respective median and mean values clearly suggest that most of the available chemicals are identified with common names (i.e. shorter), the long right-tail of labels exhibit a length comparable to IUPAC names. These longer labels are often too lengthy to be practical for use in training examples for the targeted RE task. By estimated the limit when 90\% of the chemical labels in LOTUS are at least as long as their corresponding IUPAC name, we estimated that a threshold of 60 characters effectively filters out excessively long labels.

\begin{figure}[H]
\begin{center}
  \includegraphics[width=1\textwidth]{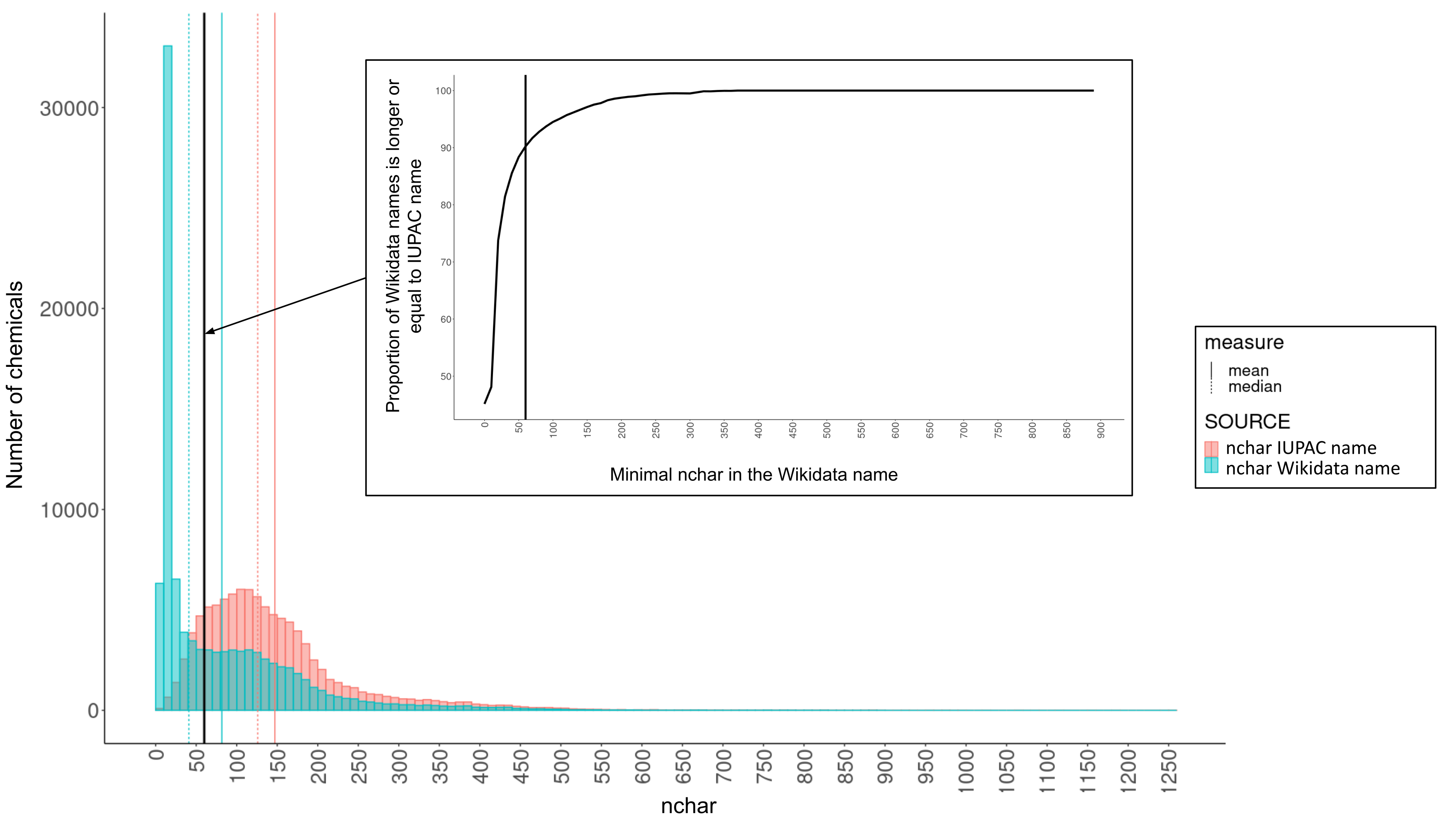}
\end{center}
\caption{Distribution of number of characters in chemical names between Wikidata labels and their corresponding IUPAC name. The integrated graph indicates the proportion of chemicals for which the length of the Wikidata label is longer or equal to the IUPAC name, by increasing number of characters in the Wikidata label. The black vertical line correspond to the chosen threshold at 60 characters.}
\label{chemicalnchar}
\end{figure}

\subsection{Diversity-sampling: alternative comparisons}
\label{sec:div-alt-strats}

\begin{figure}[H]
\begin{center}
  \includegraphics[width=1\textwidth]{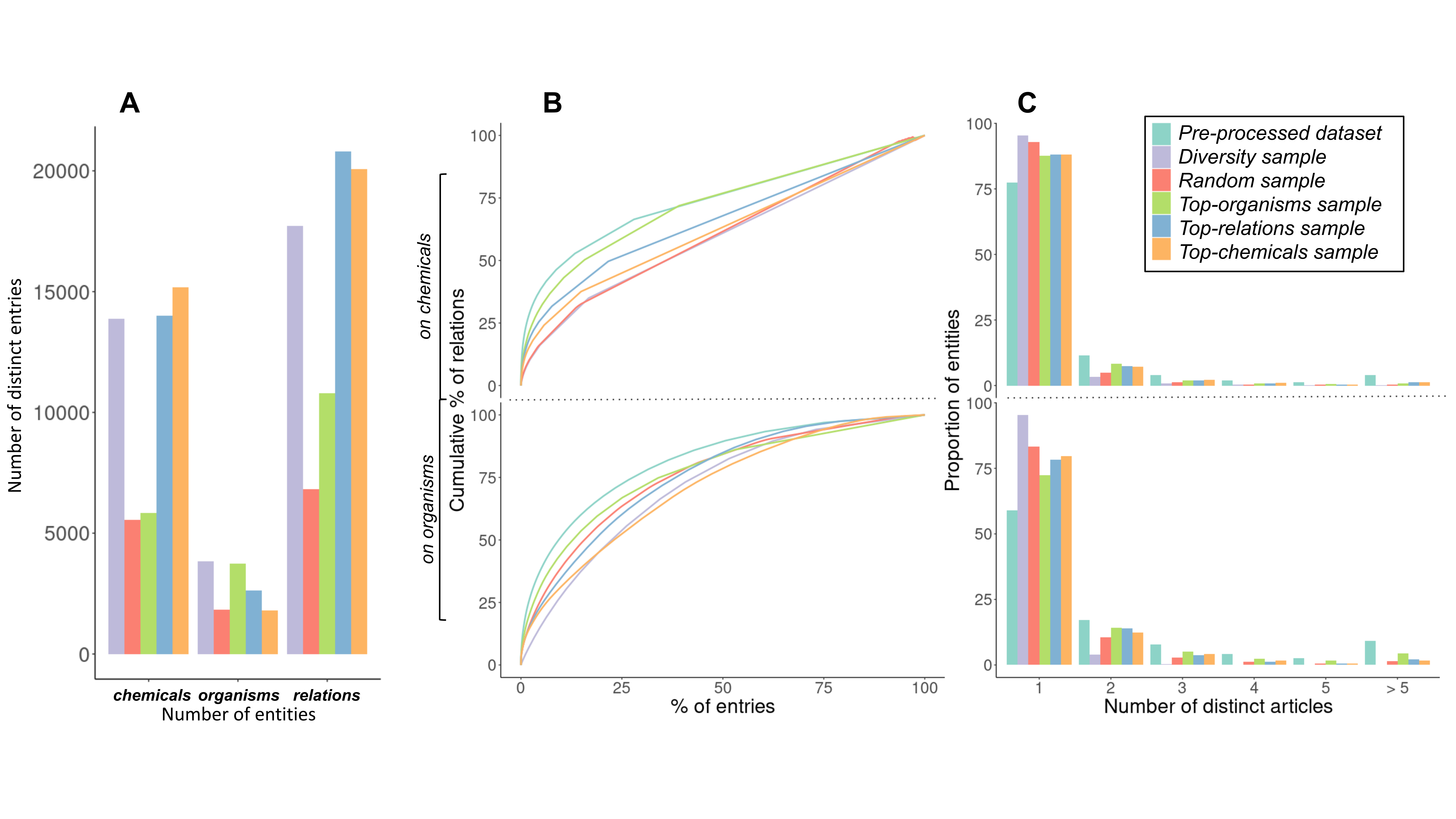}
\end{center}
\caption{Additional evaluation statistics for the diversity-sampling method compared to other potential sampling procedures. In "\textit{Top-organisms sample}", the top 500 articles with the most distinct organisms (individually) were extracted per biological kingdoms. Similarly for relations and chemicals, with "\textit{Top-relations sample}" and "\textit{Top-chemicals sample}" respectively. \textbf{A:} Statistics of the number of distinct organisms, relations, and chemicals in the \textit{Diversity} sample compared to alternative samples. All samples contain 2000 articles. \textbf{B:} Distribution of the cumulative proportion of reported relations per fraction of chemicals (top) and organisms (bottom), ordered by their contributions in the different samples. \textbf{C:} Distribution of the frequency of mention in distinct articles of chemicals (top) and organisms (bottom) in the different samples.}
\label{fig:suppEvalDatasets}
\end{figure}

The diversity-sampling strategy was evaluated against three new possible strategies for improving the diversity in the extracted set. In \textit{Top-organisms}, the top 500 articles with the most distinct organisms (individually) were extracted per biological kingdoms. Similarly done for relations and chemicals with \textit{Top-relations} and \textit{Top-chemicals}. As expected, the \textit{Top-relations} strategy led to the largest set of distinct relations (Figures \ref{fig:suppEvalDatasets}.A), followed by \textit{Top-chemicals} and the proposed diversity-sampling. However, this improvement comes at the expense of a poorer diversity in terms of organisms, but also balance in their distribution (Figures \ref{fig:suppEvalDatasets}.B and C.). Interestingly, the \textit{Top-organisms} strategy led to a smaller set of entities compared to the diversity-sampling. This highlights that in case of imbalanced distribution of entities over the sampled items (i.e. some model organisms attract more articles than non-model organisms.), the simple \textit{Top-organisms} strategy do not consider this potential redundancy, whereas its prevention is an explicit objective with the proposed approach. Overall, the evaluated metrics suggest that the diversity-sampling with the GME-sampler offers a valuable compromise between these alternative strategies.

% In total, 653,749 synonyms were extracted for 26 059 chemicals
%  for the 23,107 chemicals in the dataset

\subsection{Mismatches between standardized labels and original abstracts}
\label{sec:label-mismatches}
The 7,901 available abstracts from the literature references in the \textit{Extended} dataset were extracted using the NCBI E-utilities \texttt{efectch} service. All the organism labels available on Wikidata were directly matched on the abstracts. Using the PubChem exchange service, all the synonyms (direct synonyms of the molecule and synonyms of its stereoisomers) were extracted whether a PubChem ID was available. In total, 653,749 synonyms were extracted. A chemical entity was considered as mentioned in the abstract when there is an exact match of its name or one of its synonym in the abstract. Some chemicals, however, may also only be implicitly mentioned in an abstract. Indeed, the isolation of multiple derivatives, such as  Atroviridin A, B and C, is typical reported as “Atroviridins A-C”. Then, Atroviridin B would not be explicitly mentioned and has to be infered. All chemicals which could be part of such expressions were identified using a set of regular expressions and were treated separately to not wrongly inflate the proportion of chemicals not mentioned in the abstracts. Nonetheless, it is worth mentioning that a non-negligeable part of these \textit{multiple} chemical entities are simply not mentioned in the abstract, either explicitly or implicitly. For instance, see the original mentions of \textit{malyngamide A}\footnote{\url{https://www.wikidata.org/wiki/Q27135775}} in PMID 10924193, 11076568, and 21341718.

\subsection{Dataset curation protocol}
\label{curation-protocol}
\textbf{Biocurator: }The dataset was curated by a single curator with a PhD in microbiology and prior experiences in manual curation.\\

\textbf{Articles selection:} Articles were selected using the proposed GME-sampler, by extracting the top-200 literature references which maximise the diversity of named entities. All selected articles have a PMID, an available abstract, a title, and are available online on PubMed. No filter was applied based on the journal or the publication date.\\

\textbf{Objective:} The curator targeted the relations between organisms (\textit{head}) and their isolated natural products (\textit{tail}) in the abstracts. Only organisms and chemicals that are involved in NP relationships are extracted. For example, organisms on which the activity of a compound is tested (e.g. a pathogen like \textit{Bacillus cereus}) are not annotated. The available LOTUS annotations were always used as a starting point.\\

\textbf{Annotation of chemical entities: } All chemical entities are categorized as either singular chemical (e.g. hispaglabridin A) or chemical classes (e.g. Isoflavanoids). The nature of these entities was cross-validated with the standard ChEBI Ontology when necessary. For singular chemicals, information about their chemical class is also extracted if it is mentioned in the article. Importantly, the label of the chemical entity is annotated as it is mentioned in the abstract. To align with the original LOTUS data, Wikidata and PubChem identifiers were assigned to chemicals and classes when available. In cases of ambiguity, the curator refers to the full-text (if available) to obtain more detailed information and assign the correct standardized entity. If the entity is not found in Wikidata, a dedicated identifier in the format "\{pmid\}CHEM\{N\}" is assigned instead, e.g. "11421752CHEM1".\\

\textbf{Annotation of organism entities:} Similarly to chemicals, the name of the organism is annotated exactly as it appears in the abstract. When only the genus is determined (e.g. Plakinastrella sp.), the genus name serves as the label.\\

\textbf{Annotation of relations:} The output labels only include relations explicitly mentioned in the abstract, while relations mentioned in the full-text are excluded. The relations are annotated based on their order of appearance in the abstract. If there are more than one organism, the relations of the first organism are annotated first, followed by the relations of the other organisms in order of appearance. 

\begin{figure*}
\begin{center}
  \includegraphics[width=1\textwidth]{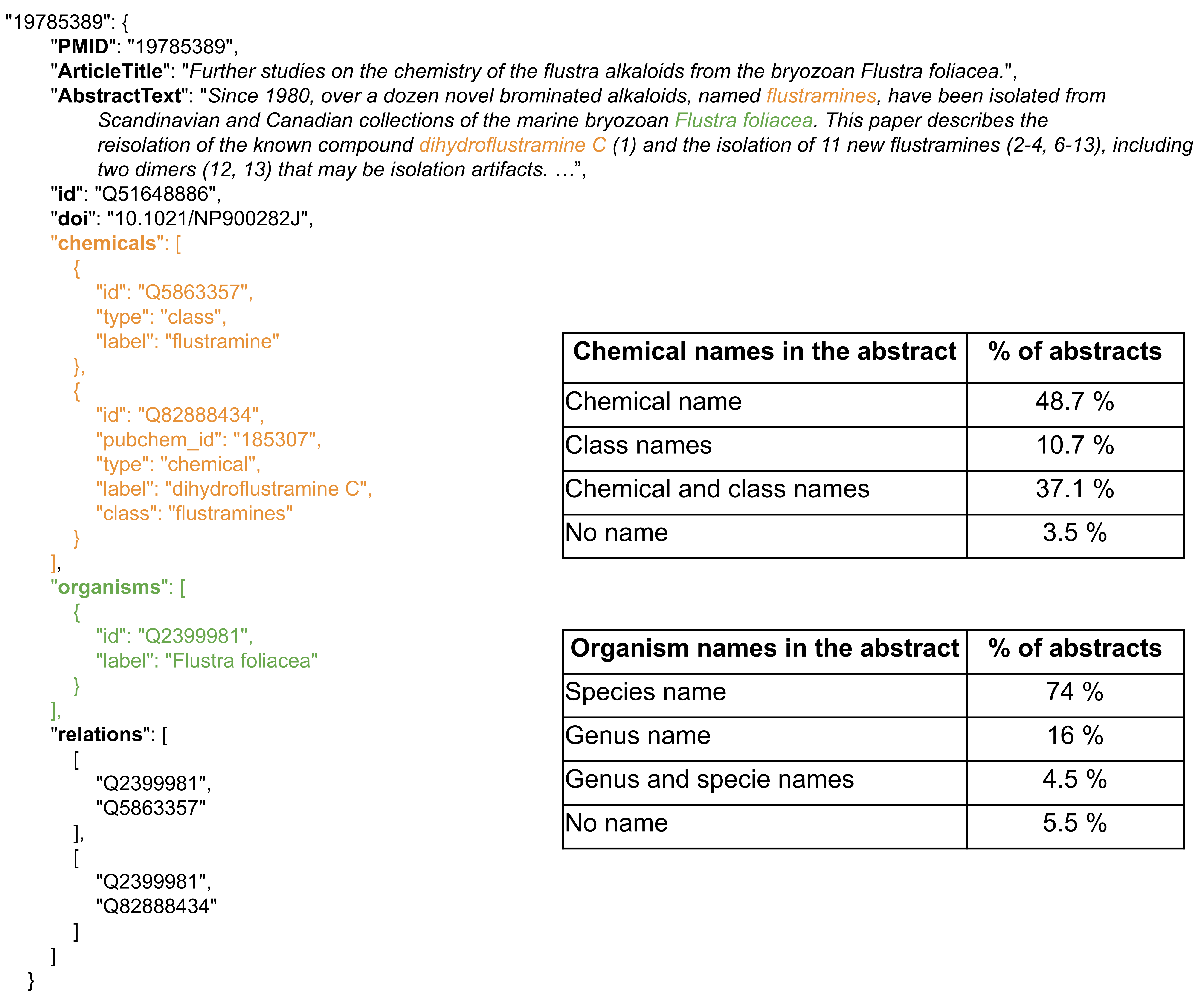}
\end{center}
\caption{An example of a curated literature reference in the evaluation dataset with supplementary statistics.}
\label{sfig:curated-dataset-eg}
\end{figure*}

\textbf{Export:} The annotations are exported in a \texttt{JSON}-format as illustrated in Figure \ref{sfig:curated-dataset-eg} along with more statistics on the annotation.

\subsection{Details of the composition of datasets}
\label{sec:curated-set-eval}

\begin{table}[H]\centering
\scriptsize
\begin{tabular}{lrrrrr}\toprule
&\textbf{\# Organisms} &\textbf{\# Chemicals} &\textbf{\# Relations} &\textbf{\# References} \\\midrule
\textbf{Original dataset} &36803 &220783 &533347 &88810 \\
\textbf{Pre-processed dataset} &14890 &56310 &102528 &32616 \\
\bottomrule
\end{tabular}
\caption{Impact of the pre-processing of the number of organisms, chemicals and relations.}\label{tab:impactPreProcessing}
\end{table}

\begin{table}[H]\centering
\scriptsize
\begin{tabular}{lrrrr}\toprule
\textbf{Kingdom} &\textbf{N} &\textbf{max $H_S(O)$ (rank)} &\textbf{max $H_S(C)$ (rank)} \\\midrule
Archaeplastida &19491 &8.73 (10512) &9.81 (10713) \\
Fungi &5023 &7.18 (2519) &9.33 (5023) \\
Metazoa &1920 &6.72 (1304) &8.33 (1920) \\
Not Attributed (Bacteria or Algae) &6666 &6.96 (2503) &8.90 (6666) \\
\bottomrule
\end{tabular}
\caption{Maximal number $N$ of literature items per kingdoms along with the value and the rank of the maximal reached entropies on organisms $H_S(O)$ and chemicals $H_S(C)$.}
\label{tab:Tab-max-entropies}
\end{table}

\begin{table}[H]\centering
\scriptsize
\begin{tabular}{lrrrrrrrrr}\toprule
&\multicolumn{4}{c}{\textbf{Organisms (\% of Max Entropy)}} &\multicolumn{4}{c}{\textbf{Chemicals (\% of Max Entropy)}} \\\midrule
\textbf{Kingdom} &\textbf{n=250} &\textbf{n=500} &\textbf{n=1000} &\textbf{n=2000} &\textbf{n=250} &\textbf{n=500} &\textbf{n=1000} &\textbf{n=2000} \\
Archaeplastida &75.5 &80.5 &86 &91.7 &76.9 &83.7 &89.6 &94.3 \\
Fungi &80.7 &89 &96.6 &99.8 &83.7 &88.1 &92.3 &95.1 \\
Metazoa &84.4 &93 &99.5 &96.1 &90 &94.6 &97.4 &100 \\
Not Attributed (Bacteria or Algae) &82.3 &90.7 &96.7 &99.9 &85.1 &89.6 &93.7 &97.2 \\
\bottomrule
\end{tabular}
\caption{Percentage of the maximal (observed) entropies $H_{\displaystyle S}(O)$ and $H_{\displaystyle S}(C)$ at different steps: 250, 500, 1000 and 2000 top ranked articles. }
\label{tab:PercentageMaxEntropy}
\end{table}

\begin{table}[H]\centering
\scriptsize
\begin{tabular}{lrrrrrr}\toprule
\textbf{Dataset} &\textbf{Part.} & \thead{\textbf{\# Relations} \\ (w. chem / w. class)} &\textbf{\# Organisms} & \thead{\textbf{\# Chemical entities} \\ (chem. / class.)} &\textbf{\# References} \\\midrule
\multirow{2}{*}{\textbf{Diversty-raw}} &train &12666 &2644 &10311 &1519* \\
&valid &1425 &301 &1211 &168* \\\midrule
\multirow{2}{*}{\textbf{Random-raw}} &train &5102 &1434 &4286 &1531* \\
&valid &657 &220 &584 &189* \\\midrule
\multirow{2}{*}{\textbf{Extended-raw}} &train &27952 &5642 &21028 &7111* \\
&valid &3355 &932 &2741 &790* \\\midrule
\multirow{2}{*}{\textbf{Full}} &train &90326 &13208 &51658 &28286 \\
&valid &1533 &484 &1288 &430 \\
\multirow{2}{*}{\textbf{Diversity-synt}} &train &11547 (10764 / 783) &2154 &(9108 / 61) &3562 \\
&valid &1197 (1096 / 101) &220 &(998 / 37) &389 \\\midrule
\multirow{2}{*}{\textbf{Random-synt }} &train &4825 (4474 / 351) &1267 &(3854 / 53 ) &3798 \\
&valid &609 (561 / 47) &190 &(507 / 22) &460 \\\midrule
\multirow{2}{*}{\textbf{Extended-synt}} &train &28614 (26373 / 2242) &5258 &(20404 / 69) &23985 \\
&valid &1444 (1332 / 112) &432 &(1122 / 37) &1254 \\
\bottomrule
\end{tabular}
\caption{Statistics about created datasets' content. For \texttt{Diversty-raw}, the top-50 articles per biological kingdoms (with an available abstract) were reserved for the evaluation set. The train/valid sets are composed of the remaining items split in 90:10. A similar split was performed on the initial 5 random samples to obtain the train/valid datasets of equivalent sizes, referred as the \texttt{Random-raw} datasets. Their count statistics are averaged over the 5 seeds. \texttt{Extended-raw} is the fusion of the \texttt{Diversty-raw} plus the 5 \texttt{Random-raw} datasets. \texttt{Full} is a dataset containing all available examples from the LOTUS snapshot, except the 200 used in the evaluation set. For synthetic datasets, the number of relations, as well as the number of distinct chemicals, is split between chemical entities and chemical classes.}
\label{tab:statsDatasets}
\end{table}

\paragraph{Curated evaluation dataset: } The composition of the curated evaluation dataset, in terms of number of distinct entities and relationships, is compared to 5 random sets of equivalent sizes. Firstly, 13 abstracts did not mention any relationships between organisms and chemicals in the curated dataset. Secondly, for the random sets, statistics were directly estimated from the LOTUS annotations. Then, they may represent an overestimate of the actual number of distinct entities, given that a manual curation could potentially eliminate some irrelevant annotations that are actually not mention in the abstracts. They should therefore be regarded as an approximate upper bound. Considering the last points, the proposed strategy for selecting the evaluation set has significantly improved the diversity, as highlighted in Table \ref{tab:testsetstats}.

\begin{table}[!htp]\centering
\scriptsize
\begin{tabular}{lrrrrr}\toprule
&\textbf{\# Organisms} &\textbf{\# Chemicals} &\textbf{\# Relations} &\textbf{\# References} \\\midrule
\textbf{eval-set (top200 diversity)} &275 &1197 ( 1092 / 105) &1488 (1297 / 191) &200 (187*) \\
\textbf{Random (200 articles)} &238 &610 &699 &200 \\
\bottomrule
\end{tabular}
\caption{Statistics of the number of organisms, chemicals, and relations in the top-200 abstracts selected and curated in the evaluation set, compared to 200 randomly selected items (statistics averaged over 5 random seeds). For the evaluation set, the number of annotated distinct chemical compounds and chemical classes are respectively indicated between parentheses. In the curated evaluation set, 13 references had no relation directly expressed in the abstract.}
\label{tab:testsetstats}
\end{table}

\subsection{Evaluation of Keyword Extraction on the SemEVAL2017 dataset}
\label{sec:kw-extraction}

\begin{table}[!htp]\centering
\scriptsize
\begin{tabular}{lrrrr}\toprule
&\textbf{TP in Top10} &\textbf{FP in Top10} &\textbf{Precision} \\\midrule
\textbf{KeyBERT (all-MiniLM-L6-v2)} &96 &904 &9.6 \\
\textbf{Vicuna-13B} &321 &669 &\textbf{32.424} \\
\bottomrule
\end{tabular}
\caption{Comparison of the performance of the prompted Vicuna-13B LLM and KeyBERT for keywords/keyphrases extraction on the \textit{SemEVAL2017} test set. The evaluation was only done on the top-10 extracted keywords for the both methods, to use the same configuration as in the experiments.}
\label{tab:evalKWExtraction}
\end{table}

The SemEVAL2017 \cite{augenstein_semeval_2017} evaluation dataset consists of 100 paragraphs, extracted from scientific publications in various domains, with on average 17.23 annotated keyphrases. While 3 subtasks are proposed in this challenge (classification and semantic relation), we only focused on the mention-level keyphrases identification. To consider similar settings as used for synthetic abstract generation, we evaluated the precision in the top-10 extracted keywords. The comparison is done by exact-match and results are presented in Table \ref{tab:evalKWExtraction}. Vicuna-13B largely outperforms the KeyBERT \cite{grootendorst2020keybert} baseline and show more than acceptable performance in zero-shot settings. KeyBERT was used with standard parameters: \texttt{keyphrase\_ngram\_range: (1,2), stop\_words: None, use\_mmr: True, diversity: 0.7} and BERT model \texttt{all-MiniLM-L6-v2} for base embeddings.

\subsection{LLM prompting for synthetic abstract generation}
\label{sec:generation-examples}
\begin{figure*}
\begin{center}
  \includegraphics[width=1\textwidth]{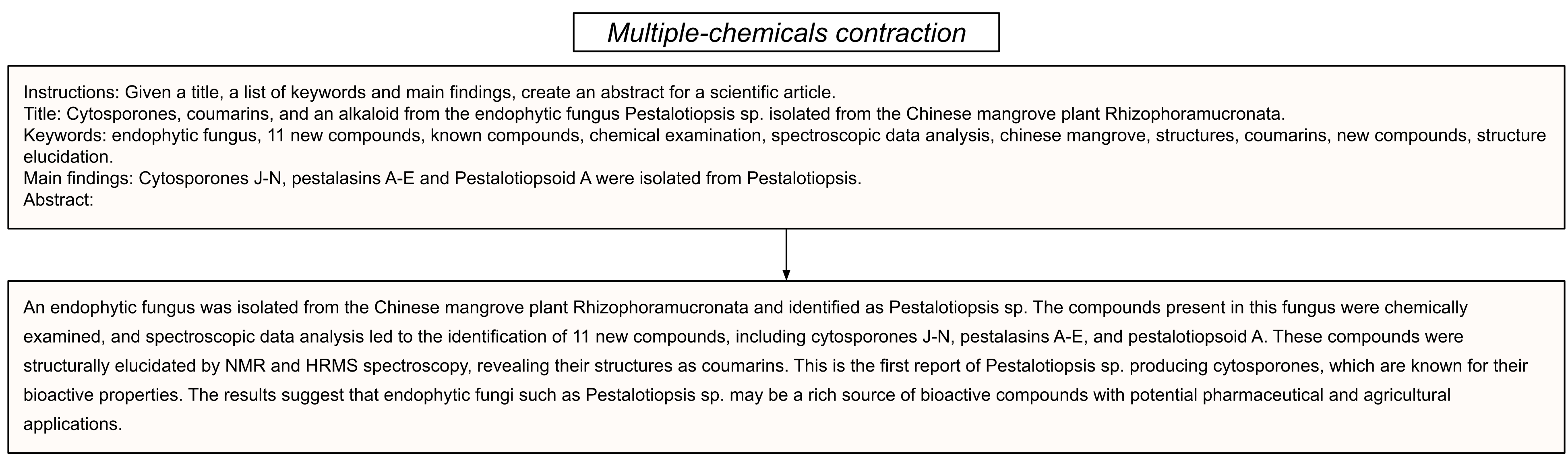}
\end{center}
\caption{An example of an abstract generation with multiple-chemical in the instruction. The original seed article is PMID 19762244. The generation was produced with $temp=0.7$. It is a training example from the \texttt{Diversity-synt} dataset.}
\label{sfig:generation-ex2}
\end{figure*}

\begin{figure*}
\begin{center}
  \includegraphics[width=1\textwidth]{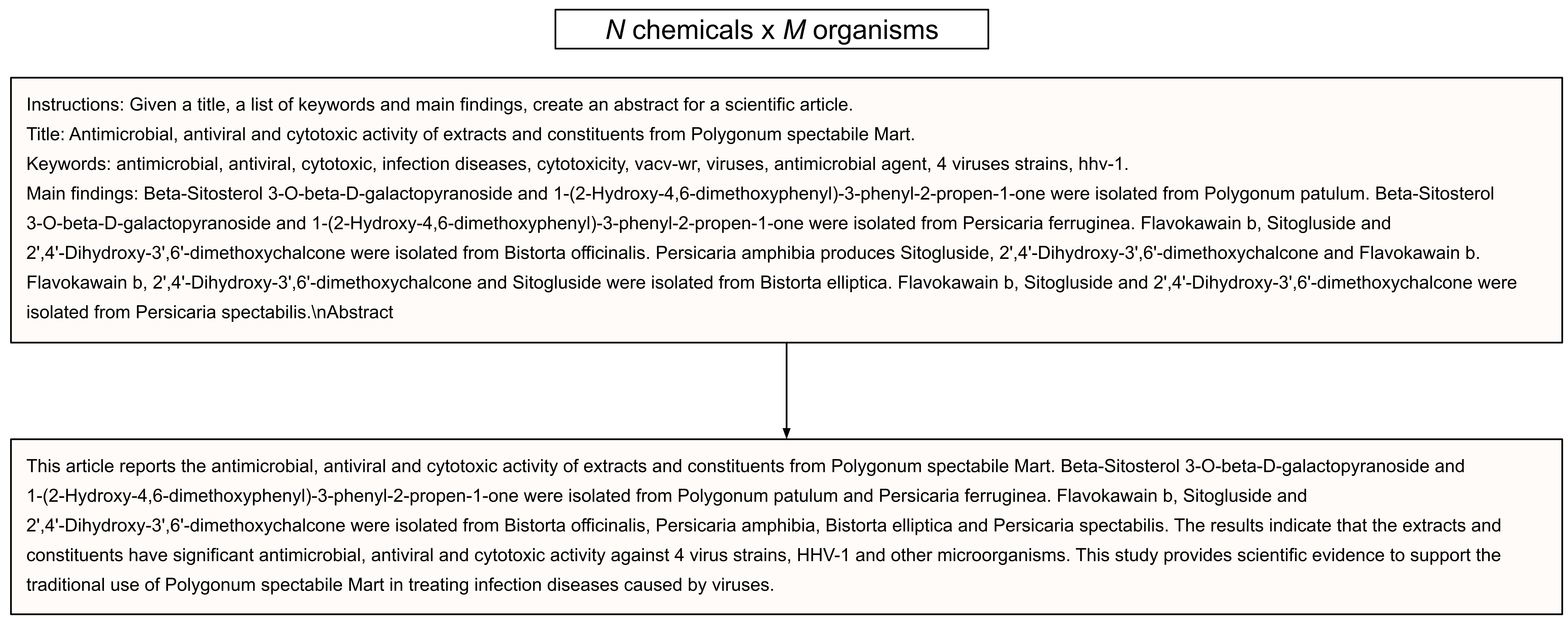}
\end{center}
\caption{An example of an abstract generation with a \texttt{N:M}-type relations. The original seed article is PMID 20382006. The generation was produced with $temp=0.7$. It is a training example from the \texttt{Diversity-synt} dataset.}
\label{sfig:generation-ex3}
\end{figure*}

The following section provides archetypal examples to illustrate the diversity engendered by the synthetic abstract generation process. Recall that each generation is calibrated with an original title, a set of keyphrases derived from the original abstract and verbalised main-findings. In the latter, 5 main transformations can be applied to improve the diversity of the generation (see Method \ref{sec:meth-synth-generation}).\\
These transformations allow for the generation of multiple alternative synthetic abstracts, which emulate different syntaxes or styles for communicating the isolation of the same set of compounds (see supplementary Figure \ref{sfig:generation-ex1}). 
\textbf{A} serves as a reference for a \textit{standard} instruction/generation. The example \textbf{B} introduces variations by reshuffling the order of the mentioned chemicals and then numbering them. In \textbf{C}, different subsets of compounds were substituted with their associated chemical families. In \textbf{A} and \textbf{B}, the expected output labels align with the verbalised main findings, e.g: "\textit{Lachnum papyraceum produces 6-Methoxymellein; Lachnum papyraceum produces 4-Chloro-6-methoxymellein, ...}". In \textbf{C}, they are substituted by the chemical classes: \textit{Lachnum papyraceum produces Coumarins; etc.}.\\
However, for "Multiple" chemicals contraction (Supplementary Figure \ref{sfig:generation-ex2}), while the synthetic text mention "cytosporones J-N, pestalasins A-E", the outputs are expected to be expanded like: "Cytosporone J, Cytosporone K, Cytosporone L, ..., Pestalasin A, Pestalasin B, ..., Pestalasin E". Verbalised relations can also exhibit a $\text{N:M}$ pattern, when multiple compounds are isolated from multiple organisms, showcasing the model creative generation abilities (Supplementary Figure \ref{sfig:generation-ex3}).\\
The generation process is subject to certain limitations and can occasionally produce inaccuracies of a similar nature to those that were intended to be mitigated. In Supplementary Figure \ref{sfig:generation-ex4}, while it is explicitly indicated in the instruction part that "Tagetes erecta produces two Flavonoids", this information does not appear in the generated abstract. Additionally, the NPs isolated from \textit{Tagetes lucida} are qualified as Flavonoids, which is a wrong assertion, i.e a hallucination. The synthetic abstracts frequently exhibit instances of hallucinations, yet, these do not significantly impair their utility for the specific task of RE, as long as they do not pertain to the expression of the relationships (see Supplementary Figure \ref{sfig:generation-ex5}).

% : (1) reordering the mention of the relations, (2) substituting a subset of chemicals with their common chemical family, (3) numbering chemicals according to the order from (1), and (4) contracting a list of multiple chemicals (e.g scupontins A-G)

\begin{figure*}
\begin{center}
  \includegraphics[width=1\textwidth]{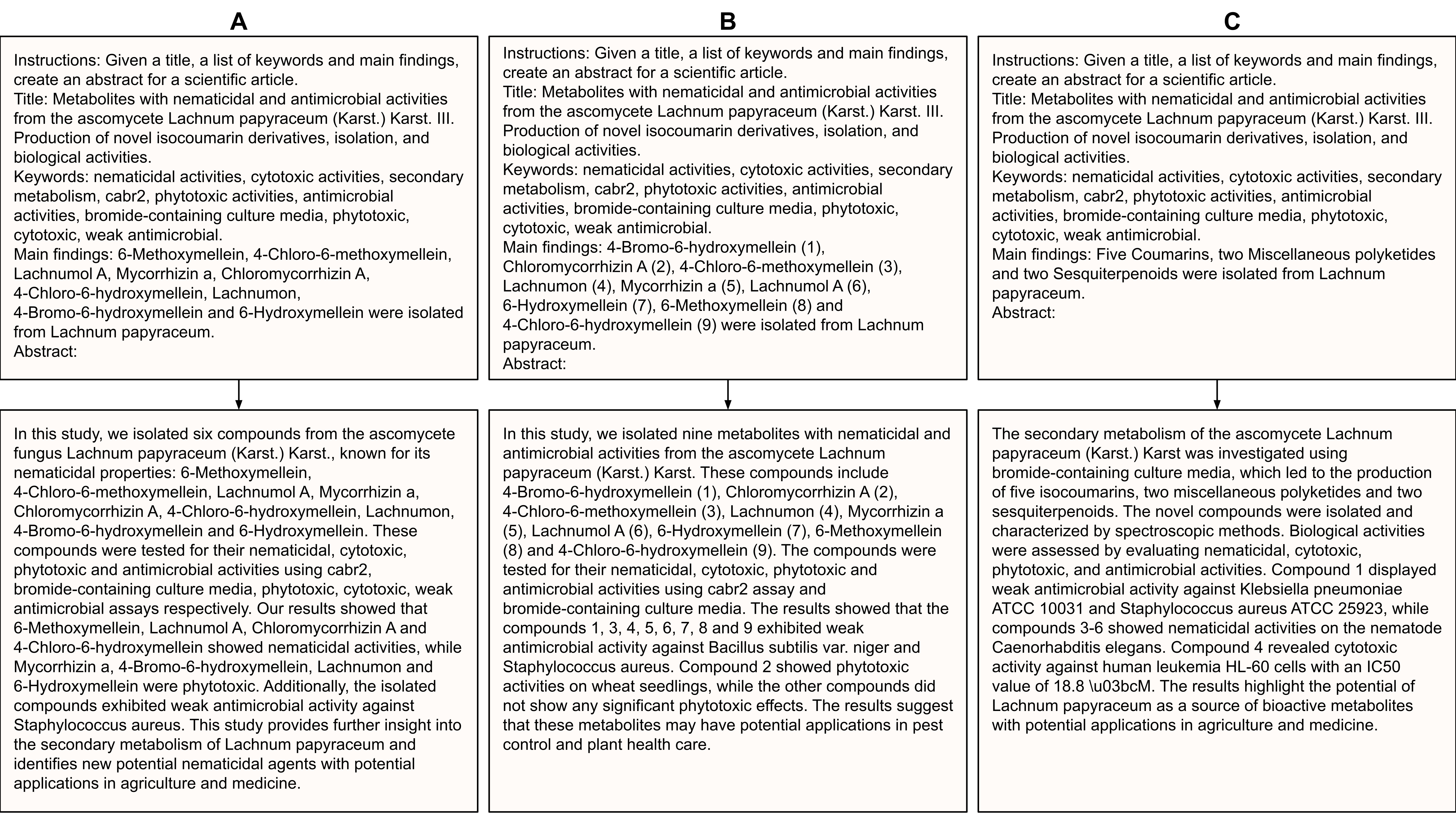}
\end{center}
\caption{3 generated abstracts from the seed article PMID: 7730162. \textbf{A} is a standard generation. In \textbf{B}, the chemicals are shuffled and numbered. In \textbf{C} chemicals were substituted with their corresponding chemical families. All generation were produced with $temp=0.7$ and are training examples from the \texttt{Diversity-synt} dataset.}
\label{sfig:generation-ex1}
\end{figure*}

\begin{figure*}
\begin{center}
  \includegraphics[width=1\textwidth]{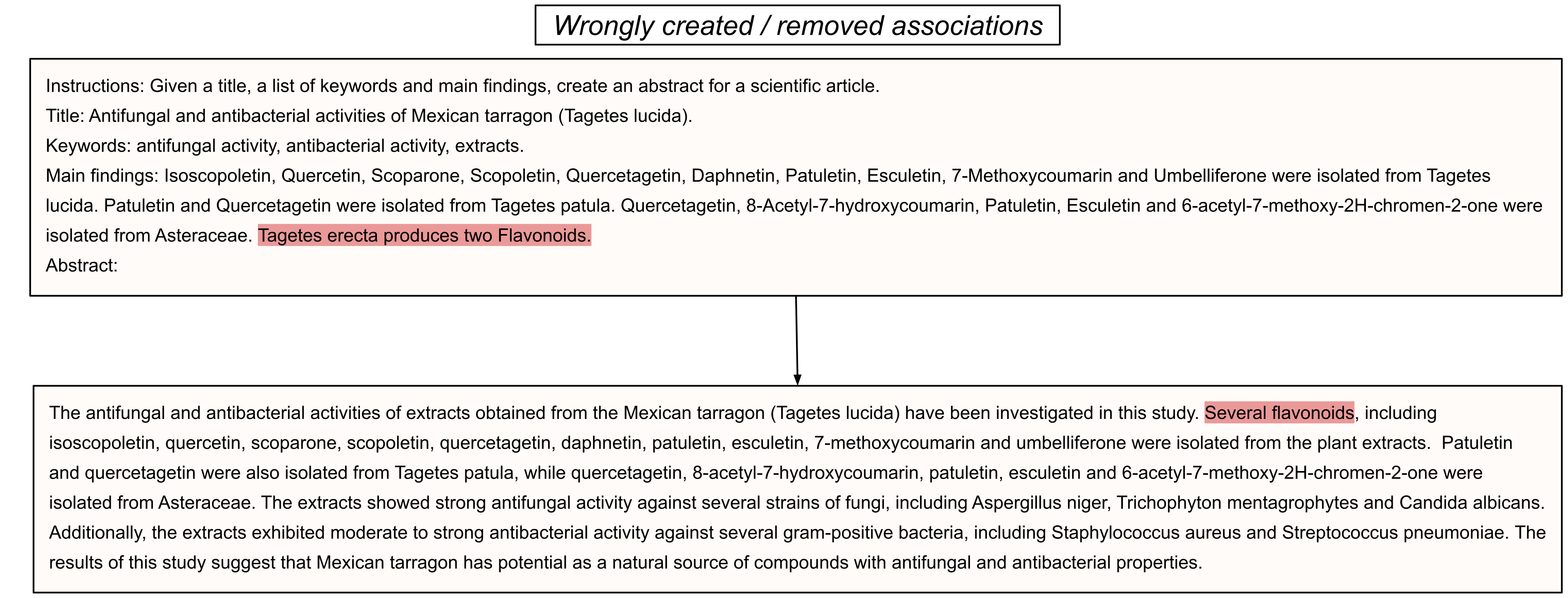}
\end{center}
\caption{An example of an abstract generation with incorrectly added and removed relations. The original seed article is PMID 19127719. The generation was produced with $temp=0.5$. It is a training example from the \texttt{Diversity-synt} dataset.}
\label{sfig:generation-ex4}
\end{figure*}

\begin{figure*}
\begin{center}
  \includegraphics[width=1\textwidth]{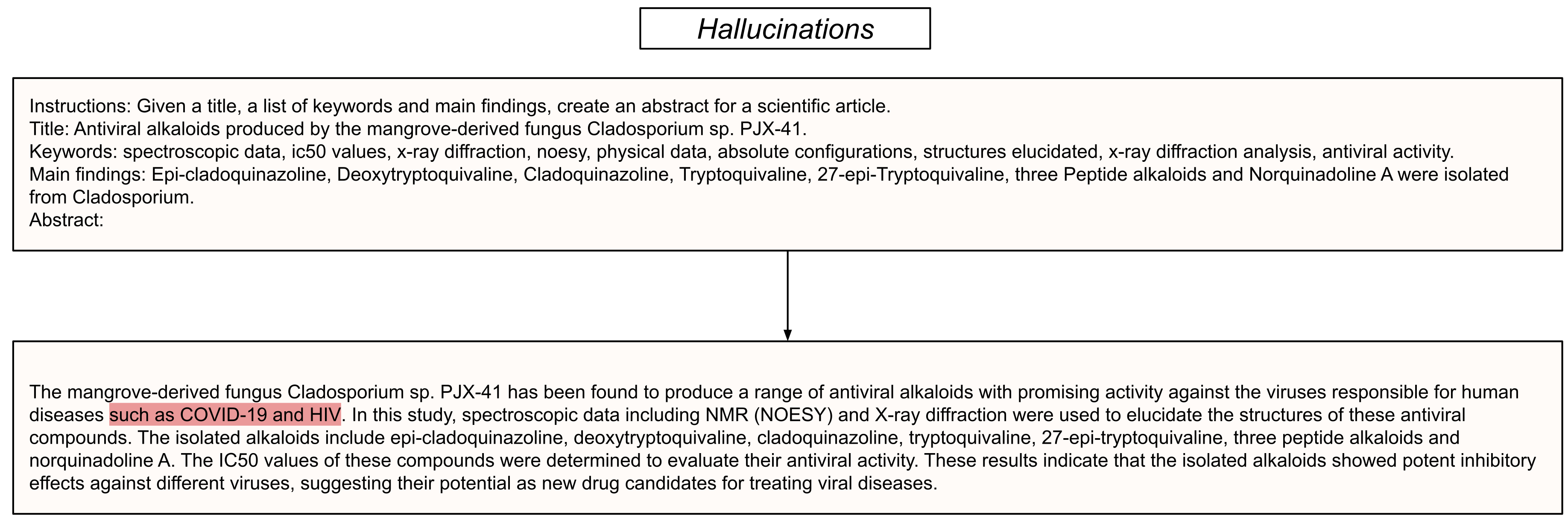}
\end{center}
\caption{An example of an abstract generation with a context hallucination. The original seed article is PMID 23758051. The generation was produced with $temp=0.5$. It is a training example from the \texttt{Diversity-synt} dataset.}
\label{sfig:generation-ex5}
\end{figure*}

\newpage
\subsection{Tokenized length of abstracts}

\begin{figure}[H]
\begin{center}
  \includegraphics[width=0.8\textwidth]{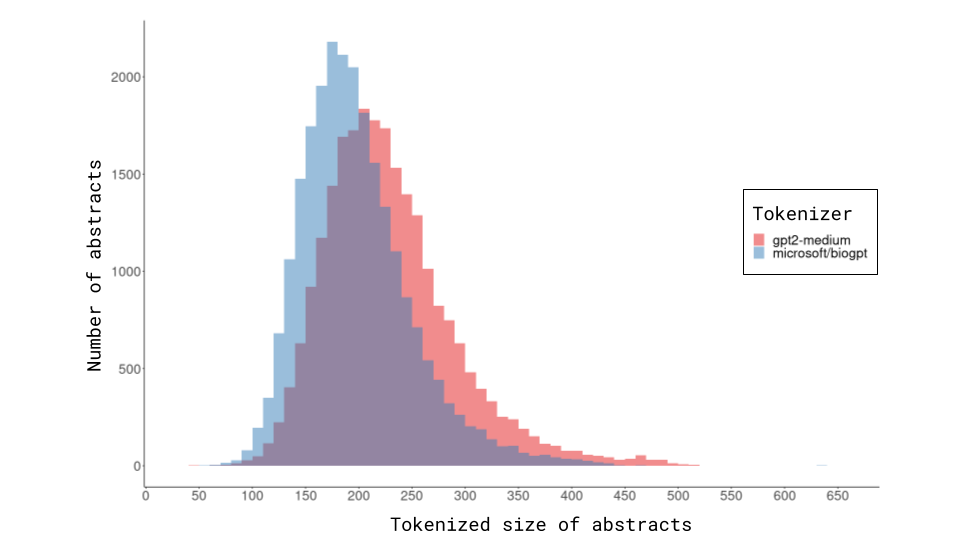}
\end{center}
\caption{Differences in size of the tokenized abstracts from the \textit{Extended-raw} dataset (23985 abstracts) between \texttt{GPT-2} and \texttt{BioGPT} tokenizers. The dedicated tokenizer of BioGPT allows for a more efficient tokenization of the abstracts.}
\label{fig:TokenizedAbstractDiff}
\end{figure}

% \input{Supplementary/SupplementaryTables/supptables}
% \input{Supplementary/SupplementaryFigures/suppfigures}

% \bibliographystylesupp{copyBmc-mathphys}
% \bibliographysupp{articlerefssupp}

\end{document}